\documentclass[preprint,12pt]{elsarticle}

\usepackage{amssymb}
\usepackage{wrapfig}
\usepackage{amsmath}
\usepackage[ruled,noend,linesnumbered]{algorithm2e}
\usepackage{booktabs}       % professional-quality tables
\usepackage{multirow, makecell}
\usepackage{subcaption}
\usepackage{hyperref}
\usepackage[table]{xcolor} % Load the xcolor package
\usepackage{caption} % For customizing captions

\journal{Artificial Intelligence}

\begin{document}

\begin{frontmatter}


\title{Learning Gradient-based Mixup with Extrapolation toward Flatter Minima for Domain Generalization}
\author[label1]{Danni Peng}
\ead{danni001@ntu.edu.sg}
\author[label2]{Sinno Jialin Pan\corref{cor1}}
\ead{sinnopan@cuhk.edu.hk}
% \fntext[label1]{}
\cortext[cor1]{Corresponding author}
\affiliation[label1]{organization={Nanyang Technological University},
            % city={},
            country={Singapore}}
\affiliation[label2]{organization={Department of Computer Science and Engineering, The Chinese University of Hong Kong},%Department and Organization
            city={Hong Kong},
            country={China}}

% \title{}


% \author{}

\begin{abstract}
To address distribution shifts between training and test data, domain generalization (DG) leverages multiple source domains to learn a model that generalizes well to unseen domains. However, existing DG methods often overfit to the source domains, partly due to the limited coverage of the expected region in feature space. Motivated by this, we propose performing mixup with data interpolation and extrapolation to cover potentially unseen regions. To prevent the detrimental effects of unconstrained extrapolation, we carefully design a policy to generate the instance weights, named \textbf{F}latness-aware \textbf{G}radient-based \textbf{Mix}up (FGMix). The policy relies on 
% gradient-based similarity 
gradient-based compatibilities to assign greater weights to instances that carry more invariant information and learn the mixup policy towards flatter minima for better generalization. On the DomainBed benchmark, we validate the efficacy of various designs of FGMix and demonstrate its superiority over other DG algorithms.
\end{abstract}

%To address the distribution shifts between training and test data, domain generalization (DG) leverages multiple source domains to learn a model that generalizes well to unseen domains. However, existing DG methods generally suffer from overfitting to the source domains, partly due to the limited coverage of the expected region in feature space. Motivated by this, we propose to perform a mixup with data interpolation and extrapolation to cover the potentially unseen regions. To prevent the detrimental effects of unconstrained extrapolation, we carefully design a policy to generate the instance weights, named \textbf{F}latness-aware \textbf{G}radient-based \textbf{Mix}up (FGMix). The policy employs a gradient-based similarity to assign greater weights to instances that carry more invariant information and learns the mixup policy towards flatter minima for better generalization. On the DomainBed benchmark, we validate the efficacy of various designs of FGMix and demonstrate its superiority over other DG algorithms.

%\item This work proposes a learnable mixup policy that enables data extrapolation and learns towards flatter minima.

%\item A gradient-based weight computation method is employed to take into account relations among examples from different domains.

%\item Extensive experiments on the DomainBed benchmark demonstrate the effectiveness of the proposed method quantitatively and qualitatively.



\end{frontmatter}


\section{Introduction}

The success of machine learning systems relies on the assumption that the training and test data are drawn from the same distribution. However, this i.i.d. assumption does not always hold in real-world applications, e.g., when the training and test data are acquired with different devices or under different conditions.
%(e.g., self-driving cars trained with day-time visionary images are required to operate also at the night-time)
When such distribution shifts occur, the systems may fail to generalize to test data if they learn to rely on spurious cues for prediction (e.g., texture or backgrounds).\

\begin{figure}[t]
\centering
\includegraphics[width=0.75\columnwidth]{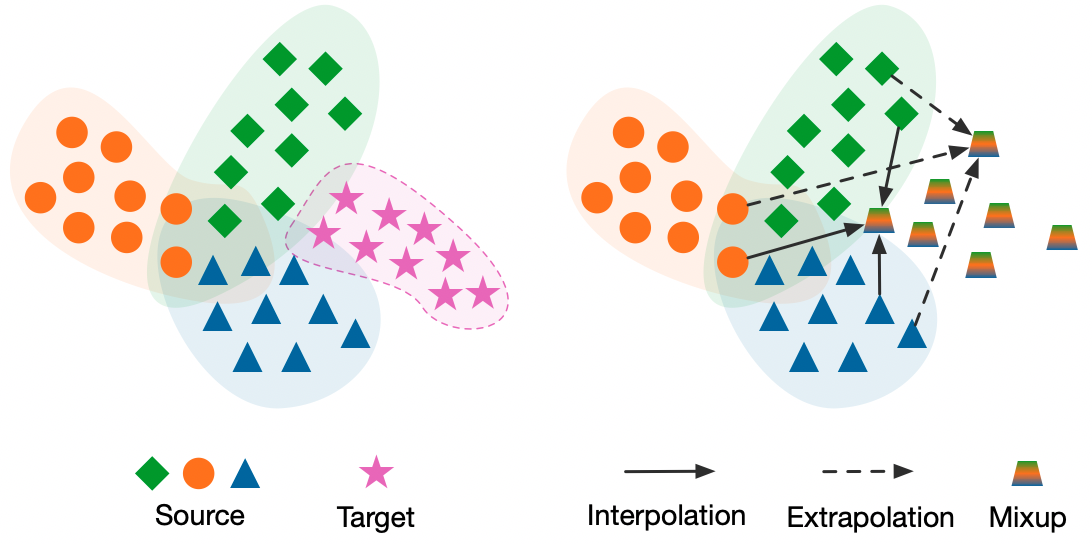}
\caption{Source domains, target domain, and mixup via interpolation/extrapolation.}
\label{fig:motivation}
\end{figure}

Domain generalization (DG) \citep{blanchard2011generalizing,muandet2013domain,li2018domain,li2017deeper} addresses this problem by leveraging data from multiple source domains to train a model that generalizes well to unseen target domains. Existing methods mainly focus on extracting invariant features from source domains or leveraging a meta-learning approach to learn a transferable model \citep{muandet2013domain,li2018domain,li2018learning,balaji2018metareg,li2019episodic}. Nevertheless, most DG methods still suffer from overfitting to the source domains.
% and poor generalization ability to the unseen target domain. 
As illustrated in Figure \ref{fig:motivation}, the left subfigure depicts some latent representations of data from multiple domains. The classifier is trained to perform well on the source domains (i.e., diamonds, circles, and triangles). If the target domain (i.e., stars) is located in a region that is not covered by the source domains, it is possible that the learned classifier will perform poorly on the target domain. Recently, mixup-based methods have been developed to address this issue \citep{zhang2018mixup,verma2019manifold,mai2021metamixup,zhou2020domain,wang2020heterogeneous}. Generally, interpolated data are used for model training such that the unseen regions within the convex hull of the source domains will also be covered, as shown by the solid arrows in the right subfigure of Figure \ref{fig:motivation}. But what if the target domain lies outside the convex hull? In that case, interpolated data is clearly insufficient, and one may need to consider using extrapolated data for model training (as shown by the dotted arrows).\

Data extrapolation is rarely involved in the existing mixup-based methods, probably due to that without a proper mixup strategy, extrapolated data may easily deviate from the expected region and become devastating to model training. Hence, unlike existing methods that simply use random weights for mixup, a carefully designed weight-generation policy is required to produce meaningful extrapolated data. To begin with, 
%Instead of randomly sampling weights from a predefined distribution, we propose to learn the policy of generating linear sample weights based on gradients. 
the weight associated with each instance involved in a linear combination should be based on its relations with other instances involved in the same combination - an idea similar to context-aware attention. Inspired by the gradient-based approach for DG \citep{li2018learning,parascandolo2020learning,mansilla2021domain,shi2021gradient}, which relies on gradient alignment for domain-invariant learning, we propose to compute the relation or 
% similarity 
compatibility score between instances based on gradients. Since the gradient 
% similarity
compatibility indicates how much information is shared between two instances from the perspective of learning, instances with a greater sum of 
% similarities 
compatibility scores with respect to all the other instances in the same combination can be considered as carrying more invariant information and hence should be assigned greater weights  \cite{eshratifar2018gradient,du2018adapting}. As a result, the mixup data will absorb a larger portion of the instances containing more invariant features. To enable extrapolation, we apply a scaling and shifting operation to the normalized weights to remove the positivity constraint. \

The generated mixup data is expected to encourage the learning of a more generalizable model. Hence, we propose to learn the two key components of our weight generation policy, i.e., the gradient-based 
% similarity 
compatibility scores and the extent of extrapolation, towards flatter minima of the classifier. A flat minimum is defined as a region in the loss surface where the loss varies slowly with changes in model parameters \citep{hochreiter1997flat}. Existing flatness-aware optimization methods \citep{hochreiter1997flat,chaudhari2019entropy,foret2020sharpness,izmailov2018averaging,guo2022stochastic} aim to search for flat minima in a given loss surface (i.e., based on the original training data). Orthogonal to these approaches, we propose flattening the loss surface that the optimizer can explore. That is, we propose to learn the weight generation policy such that the resultant loss surface based on the mixture
of the original and generated data is flatter. By flattening and widening the potential region the optimizer can explore, we avoid overfitting to the source domains and increase the likelihood of covering the target optima. As will be shown later, applying our approach together with an existing flatness-aware optimizer (one that seeks flatter minima on a given loss surface) further enhances performance. In addition to the flatness-aware objective, we further impose an auxiliary adversarial loss to constrain that the generated data does not deviate too much from a prior distribution as a mild regularization to prevent over-extrapolation.\

To summarize, we propose a \textbf{F}latness-aware \textbf{G}radient-based \textbf{Mix}up (FGMix) method that performs mixup with gradient-based instance weights and post-processing to enable extrapolation. The learnable weight generation policy is optimized to produce a flatter loss surface for better generalization and is loosely constrained to match a prior to avoid over-extrapolation. 
%In connection to a generalization bound for DG derived by \citet{cha2021swad}, we empirically show that FGMix serves to lower the bound by producing a flatter minima and reducing the divergence between source and target domains. 
Through extensive experiments, we validate the efficacy of various designs of FGMix quantitatively and qualitatively and show that FGMix achieves state-of-the-art performance on the DomainBed benchmark. \

\section{Related Work}
%\vspace{-2mm}
\paragraph{Mixup-based Methods} Mixup \citep{zhang2018mixup} is a data augmentation method that extends the training distribution by linearly interpolating random pairs of examples and labels. Incorporating mixup data for training is equivalent to minimizing the vicinal risk \citep{chapelle2000vicinal} which enables better generalization. Recently, different forms of mix-up have been developed. Cutmix \citep{yun2019cutmix} cuts out a patch from an image and switches it with another image. Remix \citep{chou2020remix} assigns greater weights to the minority class label to tackle the class imbalance issue. Manifold Mixup \citep{verma2019manifold} performs interpolations at the intermediate layers to enable smoothness in higher-level semantics. Motivated by the observation that the training set often contains few examples absent of the misleading spurious features, JM1 \cite{giannone2022just}, and SelectMix \cite{hwang2022selecmix} attempt to augment these examples by assigning greater weights to them in mixup, which are identified either through early stopping or with an auxiliary contrastive network. LISA \cite{yao2022improving} further considers the cross-domain scenario and performs both intra-class and intra-domain mixup, aiming to encourage the model to base its predictions solely on the domain-agnostic features while covering a broad spectrum of mixed-up features. Related to our work, MetaMixup \citep{mai2021metamixup} and AdaMixup \citep{guo2019mixup} learn the interpolation policy adaptively from data. The former learns by simulating pseudo-targets and pseudo-sources from the actual source domains, while the latter learns to avoid the ``manifold intrusion" issue caused by conflicts between mixup labels and original labels. Focusing on DG, MixStyle \citep{zhou2020domain} interpolates the feature statistics (known as styles) to synthesize novel domains. \citet{wang2020heterogeneous} adapt mixup to the heterogeneous setting where the label spaces are disjoint for source and target. Despite the effectiveness of various mixup methods, they primarily perform interpolation, whereas our work further explores the potential of data extrapolation to address situations where the distribution shift between source and target is significant. \

%Moreover, different from the existing data-adaptive mixup methods, we leverage gradients for generating linear weights and design a novel flatness-aware objective which directly optimizes for better generalization. 

%\vspace{-2mm}
\paragraph{Gradient-based Methods} Gradients as the update steps for SGD-based optimizers lie at the heart of deep learning algorithms. However, learning a single model for multiple tasks or distributions often runs into the problem of gradient interference, which can lead to ineffective optimization \citep{riemer2018learning}. In the context of DG, conflicting gradients often correspond to spurious domain-specific information, which can be detrimental to learning an invariant model \citep{mansilla2021domain}. The first approach to solving gradient conflicts focuses on performing some gradient surgery at each gradient step. PCGrad \citep{yu2020gradient} for multi-task learning projects a task's gradient onto the normal plane of gradients of other tasks with which it has conflicts. For DG, \citet{mansilla2021domain,parascandolo2020learning} propose masking gradient components with conflicting signs across domains. \citet{shahtalebi2021sand} further develop a smoothed-out masking method by promoting agreement among the gradient magnitudes as well. The second approach to tackle gradient conflicts typically includes gradient alignment in the learning objective. Fish \citep{shi2021gradient} explicitly optimizes the dot product between domain gradients with an efficient first-order algorithm. Fishr \citep{rame2021fishr} further enforces that the variances of gradients are matched across domains. MLDG \citep{li2018learning} employs a meta-learning approach where the meta-objective is equivalent to aligning the gradients between pseudo-source and pseudo-target domains. Different from the previous works, here we implicitly perform gradient alignment by assigning greater weights to instances whose gradients have greater overall 
% similarity 
compatibility scores to the others in the same combination. As a result, the mixup data will contain more invariant information for model learning.

%\vspace{-2mm}
\paragraph{Flatness-aware Optimization} The connection between the flatness of minima and generalization has long been established through various theories \citep{keskar2017large,hochreiter1997flat,mackay1992practical,chaudhari2019entropy}. 
% Intuitively, a flatter minimum is more robust against shifts in loss landscape between training and test data. 
In order to find a model minimizer with better generalization, existing flatness-aware methods search for flat minima in a given loss surface either by penalizing sharpness explicitly in the objective function \citep{hochreiter1997flat,chaudhari2019entropy,foret2020sharpness} or performing weight averaging to reach the flatter central region of the found minima \citep{izmailov2018averaging,guo2022stochastic}. The latter has recently been shown to obtain remarkable results on DG tasks \citep{cha2021swad,arpit2021ensemble}. Orthogonal to these approaches, we instead propose flattening the loss surface where the optimizer can explore by generating new mixup data. This also widens the attainable minimum, increasing the likelihood of covering the target domain. We later show that, when used jointly with weight averaging to reduce variance and obtain the flatter central minimum, our method yields greater improvements. 

\section{Methodology}
%\vspace{-2mm}
%\subsection{Problem Setup}
In the DG setting, suppose there are $k$ source domains $\mathcal{S}=\{\mathcal{S}_1,...,\mathcal{S}_k\}$ available for training, and we have the training data $S_i=\{(\mathbf{x}_{i,j},y_{i,j})\}_{j=1}^{|S_i|}$ drawn from the $i$-th source domain $\mathcal{S}_i$. Our goal is to learn a domain-invariant model $f:\mathbf{x}\rightarrow y$ from $S=\{S_1, ..., S_k\}$, which generalizes well to an unseen target domain $\mathcal{T}$. We assume the model is formed by two parts: a feature extractor $g_{\theta}$ and a classifier $h_{\phi}$, i.e., $f(\mathbf{x})=h_{\phi}(g_{\theta}(\mathbf{x}))$. Following \citet{berthelot2018understanding}, we refer to the data instances projected onto the latent space through $g_{\theta}$ as latent codes, i.e., the latent code of an instance $\mathbf{x}$ is $\mathbf{z}=g_{\theta}(\mathbf{x})$. The classifier is learned on top of the latent codes. Here, we only consider the homogeneous DG setting where the source and target domains share the same label space, i.e., $\mathcal{Y}_{\mathcal{S}_i}=\mathcal{Y}_\mathcal{T}, \forall i \in \{1,...,k\}$. In this case, a single classifier $h_{\phi}$ is learned and shared across all domains.

\begin{figure}[h]
\centering
\includegraphics[width=0.69\columnwidth]{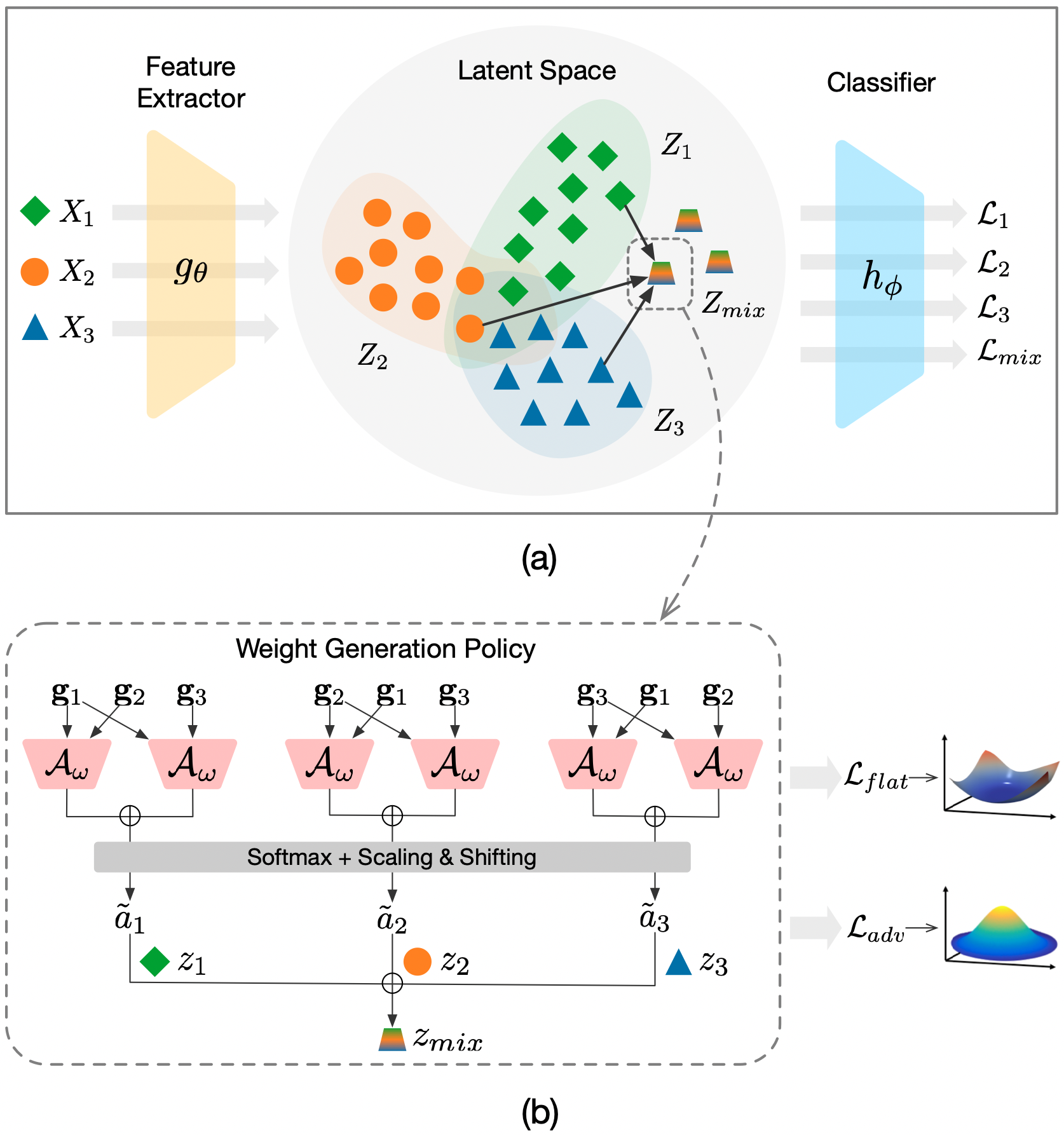}
\caption{Overview of the proposed FGMix. (a) Latent Code Augmentation: the mixup latent codes $Z_{mix}$ are used together with the original latent codes $\{Z_1, Z_2, Z_3\}$ to train the classifier $h_{\phi}$. (b) Weight Generation Policy: the weight assigned to each latent code is based on the sum of its gradient's 
% similarities 
compatibility scores w.r.t. other latent codes. A softmax and a shifting \& scaling layers are applied to enable extrapolation. The 
% similarity 
compatibility score function $\mathcal{A}_{\omega}$ and the scaling factor $\lambda$ are learned towards flatter minima (i.e., $\mathcal{L}_{flat}$) and matching the generated latent codes to a prior via adversarial training (i.e., $\mathcal{L}_{adv}$).}
\label{fig:overview}
\end{figure}

\subsection{Latent Code Augmentation}
%We follow the standard mixup implementation \cite{mai2021metamixup} to combine latent codes from multiple domains. Specifically, we generate a new latent code by linearly combining $k$ latent codes $\{\mathbf{z}_{1},...,\mathbf{z}_{k}\}$, each of which is selected from one of the $k$ source domains, and optimize the combined loss of the $k$ respective labels $\{y_{1},...,y_{k}\}$. 
We perform mixup by linearly combining latent codes of the same class from different source domains. Specifically, we sample $k$ instances $\{\mathbf{x}_1,...,\mathbf{x}_k\}$ with the same class label $y_1=...=y_k=y$, each from one of the $k$ source domains. We then feed the $k$ instances into the feature extractor $g_{\theta}$ to obtain $k$ latent codes $\{\mathbf{z}_{1},...,\mathbf{z}_{k}\}$.
% A mixup latent code is generated by linearly combining the $k$ latent codes, and the classifier $h_{\phi}$ is learned to optimize the combined loss of the $k$ labels $\{y_{1},...,y_{k}\}$ associated with the $k$ instances.
% %we generate a new latent code and label pair $(\mathbf{z}_{mix},y_{mix})$ and the respective labels . 
% %To ensure that the newly generated latent code and label pair is meaningful for the training of the classifier, these $n$ latent codes belong to the same class label, i.e., $y_{1}=...=y_{n}$. 
% Formally, 
Given the linear weights $\{a_1, ..., a_k\}$ where $\sum_{i=1}^k a_i=1$, we combine the $k$ latent codes to obtain the mixup latent code $\mathbf{z}_{mix}$ and train the classifier $h_{\phi}$ to correctly classify $\mathbf{z}_{mix}$ to the corresponding label $y$. Formally, we compute the mixup latent code $\mathbf{z}_{mix}$ and its loss $l_{mix}$ as follows:
%\vspace{-1mm}
\begin{equation}
\label{eqn:mixup}
\begin{aligned}
&\mathbf{z}_{mix}=\sum_{i=1}^k a_i\mathbf{z}_i, \ \ \ 
%&y_{mix}=y_i, \ \ \forall{i} \in \{1,...,n\},
%y_{mix}=\sum_{i=1}^k a_i y_i,
% l_{mix}=\sum_{i=1}^k a_i l(y_{i},h_{\phi}(\mathbf{z}_{mix})).\
l_{mix}=l(y,h_{\phi}(\mathbf{z}_{mix}))\
\end{aligned}
\end{equation}
%\vspace{-4mm}
%where $\sum_{i=1}^n a_i=1$. 
%where $y_i$ is one-hot representation of label. Following the standard Mixup implementation \cite{mai2021metamixup}, we perform label mixup by optimizing the expected loss for the newly generated pair $(\mathbf{z}_{mix},y_{mix})$, which is a weighted sum of losses corresponding to the individual labels:
%\begin{equation}
%	 l^{CE}(y_{mix},h_{\phi}(\mathbf{z}_{mix}))=\sum_{i=1}^k a_i l^{CE}(y_{i},h_{\phi}(\mathbf{z}_{mix})),
%\label{eqn:mixup_loss}
%\end{equation}
%where $l^{CE}(y,\hat{y})$ is the cross-entropy loss.

To allow for both interpolation and extrapolation to occur, here we do not enforce positivity constraint on $a_i$.
%\footnote{Note that positivity constraint is still applied when computing the mixup loss in \eqref{eqn:mixup_loss} as negative loss values are prohibited.}. 
Note that when $0\leq a_i\leq 1$ for all $i\in \{1,...,k\}$, the generated $\mathbf{z}_{mix}$ is an interpolation (i.e., within the convex hull formed by $\{\mathbf{z}_1,...,\mathbf{z}_k\}$), and when $a_i < 0$ for any $i \in \{1,...,k\}$, the generated $\mathbf{z}_{mix}$ is an extrapolation (i.e., outside the convex hull formed by $\{\mathbf{z}_1,...,\mathbf{z}_k\}$).\

The generated latent codes are used together with the original latent codes to train the classifier $h_{\phi}$. The feature extractor $g_{\theta}$ is also trained jointly to learn the domain-invariant, class-discriminative latent representations. Formally, suppose $n$ latent codes are generated from mixup, the model parameters $\{\theta, \phi\}$ are optimized by the following objective:
%\vspace{-1mm}
\begin{equation}
%\vspace{-1mm}
\mathop{\min}_{\theta,\phi} \frac{1}{|S|+n}\left(\sum_{i=1}^k \sum_{j=1}^{|S_i|}l(y_{i,j},h_{\phi}(\mathbf{z}_{i,j}))+\sum_{j=1}^n l_{mix,j}\right).
\label{eqn:base_loss}
\end{equation}
Training with mixup data helps prevent overfitting to the source domains and improves generalization of the learned classifier. Figure \ref{fig:overview}a shows an overview of model training with the augmented data.

\subsection{Gradient-based Weight Generation}
%\vspace{-1mm}
Unlike previous methods, which sample weights $\{a_i\}_{i=1}^k$ from a pre-defined distribution \citep{zhang2018mixup,verma2019manifold,zhou2020domain,wang2020heterogeneous}, we adopt a context-aware approach that assigns weights to instances based on their relations with other instances in the same combination. While the feature-based approach specifies 
 % \textcolor{blue}{compatibility} 
 similarity in the latent space, it does not indicate what the classifier considers as 
% \textcolor{blue}{compatible}
similar. In order for the classifier to capture the invariant information, we propose to measure the 
% \textcolor{blue}{compatibility} 
similarity between instances using gradients.\

\begin{figure}[t]
    \centering
    \begin{subfigure}[b]{0.39\textwidth}
        \centering
        \includegraphics[width=\textwidth]{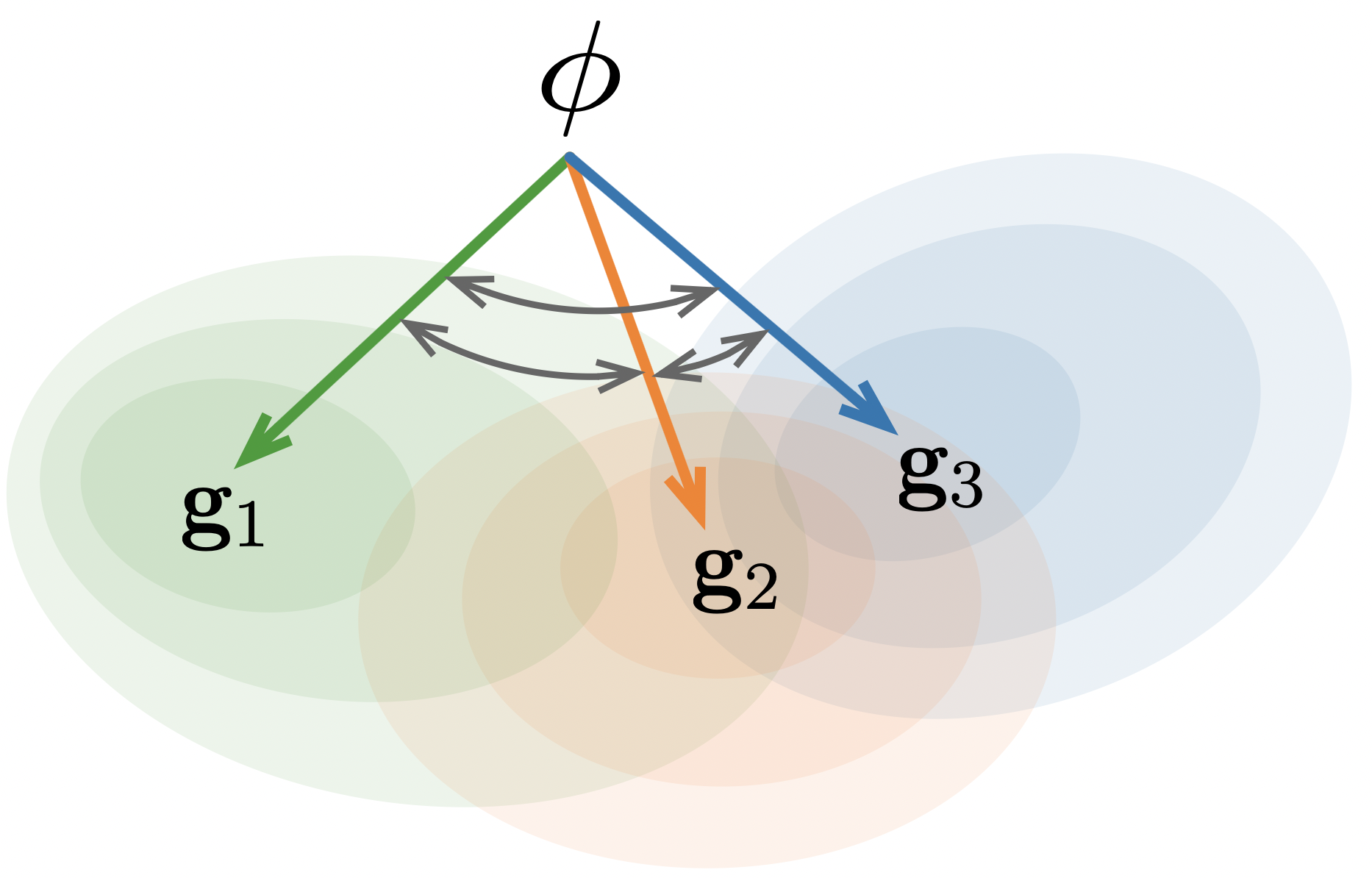}
        \caption{}
        \label{fig:gradients}
    \end{subfigure}
    \begin{subfigure}[b]{0.295\textwidth}
        \centering
        \includegraphics[width=\textwidth]{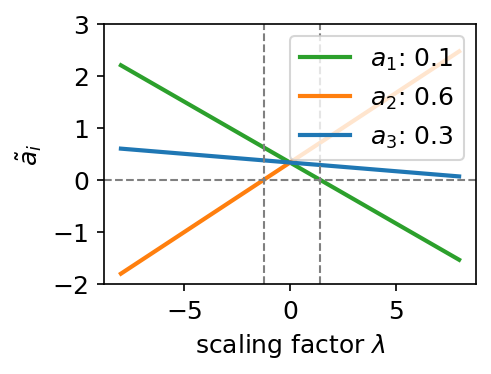}
        \caption{}
        \label{fig:sf_1}
    \end{subfigure}
    \begin{subfigure}[b]{0.295\textwidth}
        \centering
        \includegraphics[width=\textwidth]{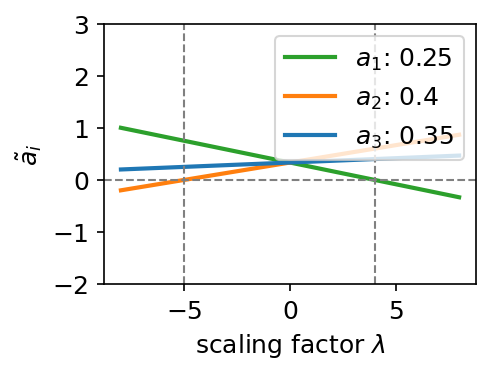}
        \caption{}
        \label{fig:sf_2}
    \end{subfigure}
    \caption{Gradients of 3 instances and the effect of the scaling factor $\lambda$ on the processed weight $\tilde{a}_i$ for different weight distributions. (a) shows gradients and the respective loss surfaces of $z_1$, $z_2$ and $z_3$ w.r.t. the classifier $\phi$. (b) and (c) illustrate the effect of $\lambda$ on $\tilde{a}_i$ for distinctive weight values (i.e., $a_1=0.1$, $a_2=0.6$, $a_3=0.3$) and similar weight values (i.e., $a_1=0.25$, $a_2=0.4$, $a_3=0.35$) respectively.}
\end{figure}

Gradient similarity quantifies the level of information sharing between instances during model learning. Leveraging gradient similarity to assign weights is not new and has been adopted by several existing works in related fields \cite{eshratifar2018gradient,du2018adapting,yu2020gradient,mansilla2021domain,parascandolo2020learning}. Here, we illustrate this concept with a toy example. Let $\mathbf{g}_i$ denote the gradient of the $i$-th instance loss $l(y_i,h_{\phi}(\mathbf{z}_i))$ w.r.t. the classifier $\phi$, i.e., $\mathbf{g}_i=\frac{\partial{l(y_i,h_{\phi}(\mathbf{z}_i))}}{\partial{\phi}}$. We consider 3 gradients $\{\mathbf{g}_1, \mathbf{g}_2, \mathbf{g}_3\}$ of instances $\{\mathbf{z}_1, \mathbf{z}_2, \mathbf{z}_3\}$ from 3 different domains as well as their respective loss surfaces, as shown in Figure \ref{fig:gradients}. Since $\mathbf{g}_2$ and $\mathbf{g}_3$ are pointing towards similar directions (i.e., $\cos{\psi_{2,3}}>0$, where $\psi_{2,3}$ is the angle between $\mathbf{g}_2$ and $\mathbf{g}_3$), taking a step along $\mathbf{g}_2$ or $\mathbf{g}_3$ will improve the classifier's performance (i.e., reducing loss) for both $\mathbf{z}_2$ and $\mathbf{z}_3$. This implies that $\mathbf{z}_2$ and $\mathbf{z}_3$ contain some shareable/invariant information as recognized by the classifier. Conversely, for $\mathbf{g}_1$ and $\mathbf{g}_3$, since they are pointing towards different directions (i.e., $\cos{\psi_{1,3}}<0$), gradient update on one will degrade the classifier's performance (i.e., increasing loss) on the other, as the level of information sharing between $\mathbf{z}_1$ and $\mathbf{z}_3$ is low. \

% In this specific example, following the direction of $\mathbf{g}_2$ seems to be the best option as it helps classify $\mathbf{z}_3$ well while does not jeopardize too much the performance on $\mathbf{z}_1$. That is, it contains the largest amount of invariant information that can be learned for the 3 involved instances. \
 
In this three-example mixup, we want to determine the relative importance of each example to assign weights. From Figure \ref{fig:gradients}, we see that $\mathbf{g}_2$ leads to a minimum where both the loss surfaces of $\mathbf{z}_1$ and $\mathbf{z}_3$ exhibit a low value, while for $\mathbf{g}_1$, it leads to a region where none of the minima of $\mathbf{z}_2$ or $\mathbf{z}_3$ overlaps with it. If the loss value indicates how well a latent code is learned by the classifier, the fact that learning $\mathbf{z}_2$ alone decreases loss for both $\mathbf{z}_1$ and $\mathbf{z}_3$ implies that $\mathbf{z}_2$ contains some sharable/invariant information acquiring which helps improve performance for both $\mathbf{z}_1$ and $\mathbf{z}_3$. Also, we observe that $\mathbf{g}_2$ possesses the greatest sum of similarities (i.e., smallest sum of angles) with respect to the other gradients (i.e., $\mathbf{g}_1$ and $\mathbf{g}_3$), while $\mathbf{g}_1$ contains the smallest sum of similarities. Thus, we can quantify the amount of invariant information carried by an instance by the sum of similarities of the instance's gradient w.r.t. all other instances, e.g., $s_i=\sum_{j\neq i}\cos{\psi_{i,j}}$, which can be used as a criterion to assign weights. When an instance exhibits high aggregate gradient similarity across domains, emphasizing it in mixup causes the gradients of the generated samples to point in directions that are consistent across domains. This consistency reduces gradient conflict and stabilizes the effective update directions, biasing optimization toward a smoother loss surface with lower curvature around the classifier parameters \cite{parascandolo2020learning,shi2021gradient}. \ref{appx:toy} illustrates an analogy between our method of quantifying invariant information and a state-of-the-art DG method. \

% Hence, it is beneficial to assign a greater weight to $\mathbf{z}_2$ and a smaller weight to $\mathbf{z}_1$ to generate a mixup encouraging the capture of invariant information.
% 
% This amount of invariant information carried by an instance can be quantified by the sum of similarities of the instance's gradient w.r.t. all the other instances, e.g., $s_i=\sum_{j\neq i}\cos{\psi_{i,j}}$, which can be used as a criterion to assign weights.

%Hence, we can see that the sum of gradient similarities of an instance w.r.t. all the other instances can be used as an indicator of the amount of invariant information the instance carries. \

%Inspired by this, we propose to generate the weight associated with each latent code based on the sum of its gradient's similarities with other latent codes. 
However, using a fixed, pre-defined measure (such as cosine similarity in the above example) to quantify similarity cannot guarantee that it yields the most suitable values for computing the mixup weights. To allow for more flexibility for the weight generation policy to generate the desired distribution, we employ a learnable function $\mathcal{A}_\omega(\cdot,\cdot)$ to measure the similarity between gradients. Specifically, we compute the sum of gradient similarities $s_i$ (also referred to as score) for the $i$-th latent code $\mathbf{z}_i$ as:
%\vspace{-1mm}
\begin{equation}
%\vspace{-1mm}
s_i=\sum_{j\neq i,j\in \{1,...,k\}}\mathcal{A}_\omega(\mathbf{g}_i,\mathbf{g}_j)	.
\label{eqn:score}
\end{equation}
In practice, we use a 2-layer MLP to model $\mathcal{A}_\omega$, where the two gradients $\mathbf{g}_i$ and $\mathbf{g}_j$ are concatenated in order (i.e., $\mathbf{g}_i$ followed by $\mathbf{g}_j$) and fed into the network, akin to similarity formulations adopted in existing metric- and relation-learning models \cite{sung2018learning,he2017neural}. Note that this similarity function is asymmetric, i.e., $\mathcal{A}_\omega(a, b)\neq \mathcal{A}_\omega(b, a)$. Hence, even when we have only 2 source domains, the 2 latent codes may be assigned different weights. Also note that since $\mathcal{A}_{\omega}$ is a learned function (with learning objectives detailed in Section 3.3), it is not guaranteed to behave exactly like a geometric similarity metric. Hence, in what follows, we refer to its output as a \textit{compatibility score} rather than a similarity score to avoid confusion. A discussion on the connection between the compatibility score function $\mathcal{A}_{\omega}$ and cosine similarity is provided in Appendix \ref{appx:corr}. Here we only require the per-example gradients w.r.t. the classifier $h_{\phi}$, which is usually shallow and close to the outputs. When applied with parallelism, e.g., using off-the-shelf vectorization functions\footnote{\url{https://pytorch.org/functorch/stable/functorch.html}}, the per-example gradient computation only incurs negligible costs.\

After that, a softmax layer is applied on top of the score $s_i$ so that the resultant weights $a_i$ sum to 1:
%\vspace{-1mm}
\begin{equation}
a_i=\frac{\exp(s_i)}{\sum_{i'=1}^k\exp(s_{i'})}.
\label{eqn:softmax}
\end{equation}
Since the softmax normalization produces $a_i\in(0,1)$, this corresponds to generating $\mathbf{z}_{mix}$ as an interpolation of $\{\mathbf{z}_1,...,\mathbf{z}_k\}$. To enable extrapolation, we further introduce a scaling and shifting procedure applied to $a_i$, which lifts the positivity constraint and allows the weight to exceed 1. 
% Generally, it serves to reduce the uncertainty in the weight distribution. 
Specifically, we introduce a scaling factor $\lambda$ and a shifting factor $\frac{\lambda-1}{k}$ to process the weight $a_i$ by:
%\vspace{-1mm}
\begin{equation}
%\vspace{-1mm}
\tilde{a}_i=\lambda a_i-	\frac{\lambda-1}{k}.
\label{eqn:scale_shift}
\end{equation}
Note that the constraint $\sum_{i=1}^k\tilde{a}_i=1$ is still fulfilled after the processing. The effect of varying the scaling factor $\lambda$ on the processed weight $\tilde{a}_i$ is shown in Figure \ref{fig:sf_1} and \ref{fig:sf_2}. We can see that the level of weight adjustment is controlled by the magnitude of $\lambda$, i.e., $|\lambda|$. For a more distinctive weight distribution (as indicated by \ref{fig:sf_1}), the effect of scaling and shifting is more significant, and extrapolation occurs at smaller $|\lambda|$ (note that extrapolation occurs when $\tilde{a}_i < 0$ for any $i$). We further allow $\lambda$ to be learnable so that the extent of extrapolation can be automatically controlled towards generating a desired mixup distribution. 
% Although the scaling–shifting expression may appear simple, the resulting extrapolation emerges from the joint optimization of $\lambda$ and $\mathcal{A}_{\omega}$, allowing the model to adaptively determine how far and in what manner extrapolation should occur to best improves generalization. 
The mixup latent code $\mathbf{z}_{mix}$ is now generated by $\mathbf{z}_{mix}=\sum_{i=1}^k \tilde{a}_i\mathbf{z}_i$.
% Note that for the mixup loss $l_{mix}$ in \eqref{eqn:mixup}, we do not apply the scaling and shifting process on the weights as negative loss values are prohibited. 
Figure \ref{fig:overview}b depicts the overall weight generation policy.

%To balance between interpolation and extrapolation, the scaling and shifting procedure is applied with a probability of $\rho \in (0,1)$, which is a hyper-parameter to be tuned for different application scenarios. \ 

\subsection{Learning the Weight Generation Policy}
%\vspace{-2mm}
To learn the two key components of our weight generation policy - the 
% similarity 
compatibility score function $\mathcal{A}_\omega$ and the scaling factor $\lambda$, we first encourage the generated mixup distribution to produce a flatter loss surface in the neighborhoods of the classifier $\phi$. This enlarges the potential region for the optimizer to explore, thereby widening the attainable source minimum and increasing the likelihood that it covers the target optimum. Generally, the flatness of the loss surface can be measured by the Hessian-based quantities \citep{keskar2017large,chaudhari2019entropy,petzka2021relative} or Monte-Carlo approximations of the loss value in the model's neighborhoods \citep{foret2020sharpness,cha2021swad}. For computational efficiency, we adopt the Monte-Carlo approach and measure flatness by the loss difference between the current classifier $\phi$ and its neighborhoods $\phi'$ within $\gamma$ distance. 
%That is, we aim to reduce the loss difference between the centre of a high-dimensional sphere and the points on its surface with radius $\gamma$. 
Formally, we optimize the weight generation policy $\omega$ and $\lambda$ with the flatness-aware loss $\mathcal{L}_{flat}$:
%\vspace{-1mm}
\begin{equation}
\begin{aligned}
&\mathop{\min}_{\omega,\lambda} \mathbb{E}_{\|\phi-\phi'\|\leq\gamma}[\triangle \mathcal{L}_{mix}^2],
\end{aligned}
\label{eqn:flat_loss}
\end{equation}
%\vspace{-1mm}
where $\triangle \mathcal{L}_{mix}=\frac{1}{n}\sum_{j=1}^nl(y_{j},h_{\phi}(\mathbf{z}_{mix,j}))-l(y_{j},h_{\phi'}(\mathbf{z}_{mix,j}))$. In practice, we approximate the expectation by sampling 100 directions from a unit sphere.
% \footnote{Appendix \ref{appx:sample_size} includes experiments to test the effects of varying the Monte-Carlo sample size.}. 

To mitigate the undesirable effects of over-extrapolation, we further impose an auxiliary adversarial loss to constrain the generated distribution so that it does not deviate substantially from the prior. Similar to \citet{li2018domain}, we match the generated distribution to a prior via adversarial training. Specifically, a discriminator $d(\cdot)$ 
%that outputs a value in the range $(0,1)$ 
is introduced to distinguish the generated latent codes from the ones sampled from the prior distribution. Our weight generation policy will learn to fool the discriminator into believing that the generated codes are from the prior. The adversarial loss $\mathcal{L}_{adv}$ is formulated as:
%\vspace{-1mm}
\begin{equation}
%\vspace{-1mm}
\mathop{\min}_{\omega,\lambda}\mathop{\max}_{d} \mathbb{E}_{\hat{\mathbf{z}}_{mix}\sim p(\hat{\mathbf{z}}_{mix})}[\log d(\hat{\mathbf{z}}_{mix})]+ \mathbb{E}_{(\mathbf{z}_1,...,\mathbf{z}_k)\sim p_{source}(\mathbf{z})}[\log(1- d(\sum_{i=1}^k \tilde{a}_i \mathbf{z}_i))],
\label{eqn:adv_loss}
\end{equation}
%\vspace{-1mm}
where $p(\hat{\mathbf{z}}_{mix})$ is a pre-defined prior distribution. In theory, we can use any arbitrary distribution for the prior. Here, we employ a Gaussian whose mean $\mu$ and variance $\sigma^2$ are computed from the source domains, i.e., $\mu=\frac{1}{|S|}\sum_{i=1}^{|S|}\mathbf{z}_i$ and $\sigma^2=\frac{1}{|S|}\sum_{i=1}^{|S|}(\mathbf{z}_i-\mu)\circ(\mathbf{z}_i-\mu)$, where $\circ$ denotes the Hadamard product  (i.e., only the diagonal entries of the covariance matrix are used). The adversarial loss is computed by sampling $\hat{\mathbf{z}}_{mix}$ from a Gaussian with $\mu$ and $\sigma^2$. In practice, since the feature extractor (and hence the latent distribution) evolves over the course of training, we compute moving averages of the mean and variance of the mini-batch to approximate $\mu$ and $\sigma^2$. In practice, we implement $d(\cdot)$ as a two-layer MLP that takes a latent code as input and outputs a probability indicating whether the code is sampled from the prior distribution or generated by the mixup module. The discriminator is trained using a standard binary cross-entropy objective to distinguish “real” latent codes drawn from the source-domain prior from generated mixup latent codes, which are treated as “fake” samples. Specifically, the discriminator is optimized by maximizing $\log d(\mathbf{z}_{real})+\log(1-d(\mathbf{z}_{fake}))$, while the generator is optimized to fool the discriminator by minimizing $\log(1-d(\mathbf{z}_{fake}))$, thereby constituting an adversarial training procedure.\

Here, adversarial training serves only as a mild constraint to prevent the generated mixup distribution from deviating substantially from the source distribution. It is less challenging to train than its more common uses, such as fine-grained image generation, which involve modeling complex mappings in high-dimensional image spaces and adhering to stricter learning criteria (e.g., realism and diversity) \cite{goodfellow2014generative}. Here, the generator (i.e., the weight generation policy) only needs to produce latent codes that mimic the prior distribution in the low-dimensional latent space \cite{makhzani2015adversarial}, and our application in FGMix further loosens the objective: as long as the generated distribution does not stray too far from the source prior, our target is fulfilled. In practice, we found that when using an adaptive optimizer such as Adam, we can achieve satisfactory results with minimal hyperparameter tuning effort.\ 

Overall, we introduce a hyper-parameter $\alpha$ to balance the contributions of $\mathcal{L}_{adv}$ and $\mathcal{L}_{flat}$. The weight generation policy is optimized towards the following combined objective:
%In other words, the discriminator $D(\cdot)$ is trained to maximize both terms by outputting larger value when the input is sampled from the prior (the first term) and smaller value when the input is generated from the policy (the second term). Conversely, our generation policy is trained to minimize the second term by fooling the discriminator to output larger value for the input it generates.\
%\textcolor{blue}{Overall, the weight generation policy is learned by optimizing the following combined loss:
\begin{equation}
\label{eqn:overall}
	\mathcal{L}_{pol}=(1-\alpha)\mathcal{L}_{flat}+\alpha\mathcal{L}_{adv}.
\end{equation}
%where $\alpha$ is a hyper-parameter to balance the two losses.}\

The weight generation policy is jointly learned with the base model $f$, i.e., we optimize \eqref{eqn:overall} simultaneously with \eqref{eqn:base_loss}. 
%, i.e., we optimize the flatness-aware loss in \eqref{eqn:flat_loss} and the adversarial loss in \eqref{eqn:adv_loss} simultaneously with the base model loss in \eqref{eqn:base_loss}. 
To ensure that the feature extractor is well-trained and produces meaningful latent codes for mixup, learning of ${\omega}$ and $\lambda$ (as well as the addition of the generated latent codes for classifier training) will only commence at the later stage of training (i.e., starting from the $\tau$-th iteration). The overall training procedure is summarized in Algorithm \ref{algo:fgmix}.

\subsection{Complexity Analysis}
%\subsection{Complexity Analysis}
Our FGMix introduces two additional modules on top of the base model, i.e., the 
% similarity 
compatibility score function $\mathcal{A}_{\omega}$ and the discriminator $d$, both adopting a lightweight architecture (i.e., a 2-layer MLP). The 
% similarity 
compatibility score function $\mathcal{A}_{\omega}$ takes in the concatenation of two gradients w.r.t. the classifier as input. Since $\mathcal{A}_{\omega}$ is asymmetric and the gradients are concatenated in order (i.e., the target gradient followed by the gradient it is compared with, as described in \eqref{eqn:score}), the 
% similarity 
compatibility score computation with $\mathcal{A}_{\omega}$ needs to be performed $k(k-1)$ times for a mixup of $k$ examples, where $k$ also refers to the number of source domains. This results in $\mathcal{O}(k^2)$ complexity for computing the mixup weights. Since, in real-world scenarios, the number of source domains $k$ is not too large (e.g., 3 to 5 source domains in our experimental datasets), this is not a significant issue. In cases with a large number of domains, we can always adopt the stochastic approach to sample $k' < k$ domains for mixup each time to control the computational cost. Once we obtain the mixup code $\mathbf{z}_{mix}$, computing the flatness-aware loss in \eqref{eqn:flat_loss} and the adversarial loss in \eqref{eqn:adv_loss} requires only constant complexity related to the size of the discriminator and the classifier, both of which are small in scale. Also note that FGMix is applied only in the later stages of training. Hence, the overall costs incurred are considered minimal as compared to the base model training.\

%\begin{figure}
% \setlength{\textfloatsep}{0pt}% Remove \textfloatsep
\begin{algorithm}[t]
\scriptsize
\SetKwInput{KwInput}{Required}                % Set the Input
\SetKwInput{KwOutput}{Output}              % set the Output
\caption{Training Procedure of FGMix}
\label{algo:fgmix}
\KwInput{Source data $S=\{S_1, ..., S_k\}$; Total number of iterations $T$; Iteration to start training and applying the weight generation policy $\tau$; Number of mixup instances generated in each iteration $n$; Learning rate of base model $\eta_1$; Learning rate of weight generation policy $\eta_2$; Neighbourhood size $\gamma$; Loss weightage $\alpha$}
\KwOutput{Learned feature extractor $g_{\theta}(\cdot)$ and classifier $h_{\phi}(\cdot)$.}
Randomly initialize all learnable parameters.\\
\For{$t \in \{1, ..., T\}$}
{
Sample a mini-batch $B=\{B_1,...,B_k\}$ from $k$ source datasets $\{S_1, ..., S_k\}$. \\
\If{$t \geq \tau$}
{
\For{$j \in {1,...,n}$}
{Sample $(\mathbf{x}_i,y_i)\in B_i$ for $i=1, ..., k$ where $y_1=...=y_k$.\\
Obtain latent code $\mathbf{z}_i=g_{\theta}(\mathbf{x}_i)$ for $i=1, ..., k$.\\
Compute linear weights $\{\tilde{a}_i\}_{i=1}^k$ by \eqref{eqn:score}-\eqref{eqn:scale_shift}.\\
Compute $\mathbf{z}_{mix,j}$ and its loss $l_{mix,j}$ by \eqref{eqn:mixup}. }
Compute base model loss $\mathcal{L}_{base}=\frac{1}{|B|+n}\left(\sum_{i=1}^k \sum_{j=1}^{|B_i|}l(y_{i,j},h_{\phi}(g_{\theta}(\mathbf{x}_{i,j})))+\sum_{j=1}^n l_{mix,j}\right)$.\\
Compute weight generation policy loss $\mathcal{L}_{pol}=(1-\alpha)\mathbb{E}_{\|\phi-\phi'\|\leq\gamma}[\triangle \mathcal{L}_{mix}^2]+\alpha[\frac{1}{n}\sum_{j=1}^n\log(1- d(\mathbf{z}_{mix,j}))]$.\\
Compute discriminator loss $\mathcal{L}_d=-\alpha[\frac{1}{n}\sum_{j=1}^n(\log d(\hat{\mathbf{z}}_{mix,j}) + \log(1- d(\mathbf{z}_{mix,j})))]$.\\
Update weight generation policy and discriminator $\omega \leftarrow \omega - \eta_2\nabla_{\omega}\mathcal{L}_{pol}$, $\lambda \leftarrow \lambda - \eta_2\nabla_{\lambda}\mathcal{L}_{pol}$, $d \leftarrow d - \eta_2\nabla_{d}\mathcal{L}_{d}$.\\
}
\Else
{Compute base model loss $\mathcal{L}_{base}=\frac{1}{|B|}\sum_{i=1}^k \sum_{j=1}^{|B_i|}l(y_{i,j},h_{\phi}(g_{\theta}(\mathbf{x}_{i,j})))$.\\}
Update base model $\theta \leftarrow \theta - \eta_1\nabla_{\theta}\mathcal{L}_{base}$, $\phi \leftarrow \phi - \eta_1\nabla_{\phi}\mathcal{L}_{base}$.
}
\end{algorithm}
% \setlength{\textfloatsep}{8pt}% Remove \textfloatsep
%\vspace{-10mm}
%\end{figure}
%\setlength{\textfloatsep}{5pt}% Remove \textfloatsep
%\vspace{-5mm}
\clearpage

\section{Experiments}
\label{sec:experiments}
%\vspace{-2mm}
\subsection{Experimental Setup}
%\subsubsection{Dataset Details} 
We conduct experiments mainly on DomainBed \citep{gulrajani2020search}, a recently introduced testbed that provides a unified evaluation procedure for DG algorithms. Following the previous DG studies \citep{arpit2021ensemble,cha2021swad}, we focus on five real-world benchmark datasets available on DomainBed: \texttt{PACS} \citep{li2017deeper} (4 domains, 7 classes, and 9,991 images), \texttt{VLCS} \citep{fang2013unbiased} (4 domains, 5 classes, and 10,729 images), \texttt{OfficeHome} \citep{venkateswara2017deep} (4 domains, 65 classes, and 15,588 images), \texttt{TerraIncognita}
\citep{beery2018recognition} (4 domains, 10 classes, and 24,788 images) and \texttt{DomainNet} \citep{peng2019moment} (6 domains, 345 classes, and 586,575 images).

%\vspace{-2mm}
%\subsubsection{Implementation Details} 
For a fair comparison, we follow the experimental settings by \citet{gulrajani2020search}, including data splits (20\% data are reserved for validation for each training domain), hyper-parameter search (a search distribution is pre-defined for each hyper-parameter), number of iterations (default to be 15,000 for \texttt{DomainNet} and 5,000 for the other datasets), image augmentation (cropping, resizing, horizontal ﬂips, color jitter, grayscaling, normalization, etc.) and the base model backbone (ResNet-50 \citep{he2016deep} pre-trained on ImageNet as initialization).\
%For \texttt{DomainNet} dataset, we follow the recent studies by \citet{arpit2021ensemble} and \citet{cha2021s

For our proposed FGMix, the number of instances $k$ used for mixup is set to the number of source domains in each dataset ($k=3$ for \texttt{PACS}, \texttt{VLCS}, \texttt{OfficeHome}, and \texttt{TerraInc}; $k=5$ for \texttt{DomainNet}), with one instance sampled from each source domain. We set the iteration $\tau$ to start training and applying weight generation for mixup as 9,000 for \texttt{DomainNet} and 3,000 for the other datasets and set the size of mixup data  $n$ generated in each iteration as 32, same as the batch size for each source domain\footnote{Note that since \texttt{DomainNet} has large label size (345 classes in total), applying the default random batch sampler in DomainBed cannot ensure that there are enough same-class examples in each batch for mixup. Hence, we modify the batch sampler by first sampling $m$ classes ($m \ll$ batch size) and then drawing examples from the $m$ classes to form the mini-batch for each source domain.}. We tune the learning rate of weight generation policy and the discriminator in \{1e-4, 1e-3, 1e-2\}, the initialization of the scaling factor $\lambda$ in \{1, 3, 5, 8\}\footnote{Note that $\lambda=1$ is equivalent to having no scaling \& shifting effect.}, the neighborhood size $\gamma$ in $10^{\text{Uniform}(0,2)}$, and the loss weightage $\alpha$ in \{0.01, 0.03, 0.1, 0.3\} on the training domain validation sets. All experiments are repeated for 5 trials with different random seeds\footnote{Experiments are conducted on NVIDIA A100 with 40GB memory.}. Regarding the network architecture, we use a 2-layer MLP for the 
% similarity 
compatibility score function $\mathcal{A}_{\omega}$ accompanied with $\tanh$ activations. The hidden sizes for both layers are set to 64. The discriminator $d$ is also a 2-layer MLP with a hidden size 64. 
% To apply SWA, we follow \citet{arpit2021ensemble} which suggests skipping the first few iterations and start applying weight averaging from the 100-th iteration.

%\vspace{-1mm}
\subsection{Baseline Methods and Hyperparameter Settings}
%\vspace{-1mm}
We select several baselines related to our method for overall comparison on DomainBed\footnote{{We note that ERM++\cite{teterwak2023erm++} is also evaluated on DomainBed but employs several modifications to the standard setup. Hence, it is not included here for a fair comparison.}}: (1) naive baseline without any DG strategy: \textbf{ERM} \citep{vapnick1998statistical} (empirical risk minimization); (2) mixup-based methods: \textbf{Mixup} \citep{wang2020heterogeneous} (mixup at the input level), \textbf{Manifold Mixup} \citep{verma2019manifold} (mixup at the feature level), \textbf{MixStyle} \citep{zhou2020domain} (mixup of feature statistics) and \textbf{MetaMixup} \citep{mai2021metamixup} (meta-learn the interpolation policy); (3) gradient-based methods: \textbf{PCGrad} \citep{yu2020gradient} (project gradients onto the normal plane of conflicting gradients), \textbf{AND-Mask} \citep{parascandolo2020learning} (mask out conflicting gradient components), \textbf{Fish} \citep{shi2021gradient} (optimize for gradient alignment) and \textbf{Fishr} \citep{rame2021fishr} (optimize for gradient covariance alignment); (4) augmentation methods: \textbf{L2A-OT} \citep{zhou2020learning} (generate pseudo-source by enlarging divergence to the source domains), 
%\textbf{Style Neophile} \citep{kang2022style} (generate pseudo-source by random jittering and careful selection), 
\textbf{CNSN} \citep{tang2021crossnorm} (exchange and normalize instances' styles), \textbf{DDG} \citep{zhang2022towards} (disentangle and swap instances' variation factors) (5) current SOTA on DomainBed: \textbf{SagNet} \citep{nam2021reducing} (reduce style bias), \textbf{SelfReg} \citep{kim2021selfreg} (self-supervised contrastive regularization) and \textbf{CORAL} \citep{sun2016deep} (correlation alignment).

%\subsection{Hyperparameters Settings}
%\label{appx:hyperparams}
For our FGMix and all the reproduced algorithms (except for DDG, which will be detailed later), we train the base model by setting the batch size as 32 (due to constraints in computational resources) and tuning the dropout rate in \{0, 0.1, 0.5\}, learning rate in \{1e-5, 3e-5, 5e-5\} and weight decay in \{1e-4, 1e-6\}, following \citet{cha2021swad}.\

We reproduced 7 baselines (i.e., Manifold Mixup \citep{verma2019manifold}, MixStyle \citep{zhou2020domain}, MetaMixup \citep{mai2021metamixup}, PCGrad \citep{yu2020gradient}, L2A-OT \citep{zhou2020learning}, 
%Style Neophile \citep{kang2022style}, 
CNSN \citep{tang2021crossnorm} and DDG \citep{zhang2022towards}) due to their lack of results on DomainBed benchmark \citep{gulrajani2020search}. We directly adopt the official implementations released by the respective authors in the DomainBed environment (refer to the appendix of \citet{gulrajani2020search} for how to incorporate new algorithms into DomainBed). \

For Manifold Mixup, we set $\alpha=0.2$ for the beta distribution from which the interpolation constant is sampled. For MixStyle, we set $\alpha=0.1$ for the Beta distribution and $p=0.5$ for the probability of applying MixStyle. As recommended, we insert the MixStyle layer after the 1st and 2nd residual blocks. For MetaMixup, we set the learning rate for the interpolation constant to 1e-3 (similar to our weight generation policy). PCGrad is free of additional hyperparameters. For L2A-OT, we use ResNet-50 as both the classifier and the domain discriminator to compute the domain divergence. Following the suggestions by the authors, we search $
\lambda_{\text{Domain}}$ in \{0.5, 1, 2\}, $
\lambda_{\text{Cycle}}$ in \{10, 20\}\ and $
\lambda_{\text{CE}}$ in \{1\}. For CNSN, we apply both CrossNorm (2-instance mode) and SelfNorm, inserting them at the end of each residual block in ResNet-50. We set the number of active CrossNorm layers to 1 and the probability of applying CrossNorm to 0.5 to avoid over-augmentation. For DDG, a 2-stage procedure is adopted, where the generator is pre-trained as part of a GAN in the first stage and then applied (and further updated) in the second stage. Following the official implementation, the batch size is set to 2, and the number of training steps is enlarged to 25,000 for the first stage and 10,000 for the second stage. Due to the 2-stage training and the larger number of training steps, DDG takes much longer to train than other baselines, making it less competitive.

To apply SWA, we follow \citet{arpit2021ensemble}, which suggests skipping the first few iterations and starting to apply weight averaging from the 100th iteration.

\subsection{Overall Comparison}
\begin{table}[t]
%\vspace{-8mm}
\caption{Overall comparison of selected algorithms on five datasets. Model selection is based on \textbf{training domain validation}. The reported result for each dataset is the average performance across all test domains for that dataset. Most of the results are obtained directly from the literature, except for those denoted with \textsuperscript{\textdagger}, which are from our reproduction on DomainBed. We also report the time cost (in ms) per training iteration, averaged over 5000 iterations. The time reported is for datasets with 3 source domains, and thus uses $k=3$ instances for mixup each time.
%, and those denoted with \textsuperscript{\textdagger\textdagger}, which are from \citet{arpit2021ensemble} using 15,000 iterations for \texttt{DomainNet}. 
}
%\vspace{-2mm}

\setlength{\tabcolsep}{2.2pt}
\label{tbl:overall}
\scriptsize
\centering
\begin{tabular}{c l | c c c c c | c | c}
\toprule
& \textbf{Algorithm} & \texttt{PACS} & \texttt{VLCS} & \texttt{OfficeHome} & \texttt{TerraInc} & \texttt{DomainNet} & Avg. & Time\\
\midrule
& ERM \citep{vapnick1998statistical} & 85.5\tiny$\pm$0.2 & 77.5\tiny$\pm$0.4 & 66.5\tiny$\pm$0.3 & 46.1\tiny$\pm$1.8 & 40.9\tiny$\pm$0.1 & 63.3 & 284.8 \\
\midrule
\multirowcell{4}{\textit{Mixup-based} \\ \textit{Methods}} & Mixup \citep{wang2020heterogeneous} & 84.6\tiny$\pm$0.6 & 77.4\tiny$\pm$0.6 & 68.1\tiny$\pm$0.3 & 47.9\tiny$\pm$0.8 & 39.2\tiny$\pm$0.1 & 63.4 & 285.6\\
& Manifold Mixup\textsuperscript{\textdagger} \citep{verma2019manifold} & 86.2\tiny$\pm$0.6 & 76.7\tiny$\pm$1.1 & 67.3\tiny$\pm$1.0 & 48.8\tiny$\pm$2.1 & 41.2\tiny$\pm$0.3 & 64.0 & 444.5\\
& MixStyle\textsuperscript{\textdagger} \citep{zhou2020domain} & 85.3\tiny$\pm$1.9 & 77.4\tiny$\pm$0.8 & 67.3\tiny$\pm$0.9 & 46.8\tiny$\pm$1.1 & 40.9\tiny$\pm$0.2 & 63.5 & 290.2\\
& MetaMixup\textsuperscript{\textdagger} \citep{mai2021metamixup} & 85.2\tiny$\pm$1.2 & 77.9\tiny$\pm$0.8 & 68.2\tiny$\pm$0.7 & 47.1\tiny$\pm$1.8 & 41.8\tiny$\pm$0.3 & 64.0 & 750.7\\
\midrule
\multirowcell{4}{\textit{Gradient-based} \\ \textit{Methods}} & PCGrad\textsuperscript{\textdagger} \citep{yu2020gradient} & 85.0\tiny$\pm$0.9
& 77.5\tiny$\pm$0.8 & 65.5\tiny$\pm$0.9 & 48.3\tiny$\pm$2.3 & 41.1\tiny$\pm$0.2 & 63.5 & 310.0\\
& AND-Mask \citep{parascandolo2020learning} & 84.4\tiny$\pm$0.9 & 78.1\tiny$\pm$0.9 & 65.6\tiny$\pm$0.4 & 44.6\tiny$\pm$0.3 & 37.2\tiny$\pm$0.6 & 62.0 & 319.4\\
& Fish \citep{shi2021gradient} & 85.5\tiny$\pm$0.3 & 77.8\tiny$\pm$0.3 & 68.6\tiny$\pm$0.4 & 45.1\tiny$\pm$1.3 & \textbf{42.7}\tiny$\pm$0.2 & 63.9 & 671.7 \\
& Fishr \citep{rame2021fishr} & 85.5\tiny$\pm$0.4 & 77.8\tiny$\pm$0.1 & 67.8\tiny$\pm$0.1 & 47.4\tiny$\pm$1.6 & 41.7\tiny$\pm$0.0 & 64.0 & 703.3\\
\midrule
\multirowcell{3}{\textit{Augmentation} \\ \textit{Methods}} & L2A-OT\textsuperscript{\textdagger} \citep{zhou2020learning} & 85.8\tiny$\pm$1.9 & 77.4\tiny$\pm$0.9 & 68.1\tiny$\pm$1.6 & 48.6\tiny$\pm$2.1 & 40.2\tiny$\pm$0.5 & 64.0 & 622.7\\
& CNSN\textsuperscript{\textdagger} \citep{tang2021crossnorm} & 85.6\tiny$\pm$0.9 & 77.1\tiny$\pm$1.0 & 67.3\tiny$\pm$1.2 & 48.4\tiny$\pm$1.6 & 40.7\tiny$\pm$0.4 & 63.8 & 412.1\\
& DDG\textsuperscript{\textdagger} \citep{zhang2022towards} & 85.3\tiny$\pm$1.6 & 76.8\tiny$\pm$1.2 & 68.1\tiny$\pm$1.0 & 47.7\tiny$\pm$2.1 & 40.0\tiny$\pm$0.5 & 63.6 & 653.2\\
\midrule
\multirowcell{3}{\textit{DomainBed} \\ \textit{SOTA}} & SelfReg \citep{kim2021selfreg} & 85.6\tiny$\pm$0.4 & 77.8\tiny$\pm$0.9 & 67.9\tiny$\pm$0.7 & 47.0\tiny$\pm$0.3 & 41.5\tiny$\pm$0.2 & 64.0 & 320.1\\
& SagNet \citep{nam2021reducing} & 86.3\tiny$\pm$0.2 & 77.8\tiny$\pm$0.5 & 68.1\tiny$\pm$0.1 & 48.6\tiny$\pm$1.0 & 40.3\tiny$\pm$0.1 & 64.2 & 671.7 \\
& CORAL \citep{sun2016deep} & 86.2\tiny$\pm$0.3 & \textbf{78.8}\tiny$\pm$0.6 & 68.7\tiny$\pm$0.3 & 47.6\tiny$\pm$1.0 & 41.5\tiny$\pm$0.1 & 64.6 & 305.8 \\ 
\midrule
& FGMix (ours) & \textbf{86.8}\tiny$\pm$1.2 & 78.1\tiny$\pm$0.9 & \textbf{69.0}\tiny$\pm$1.3 & \textbf{49.2}\tiny$\pm$1.4 & 42.1\tiny$\pm$0.5 & \textbf{65.0} & 312.7\\
\midrule
\multicolumn{4}{c}{\textit{Combined with flatness-aware optimizer SWA \citep{izmailov2018averaging,arpit2021ensemble}}}  \\
\midrule
& ERM + SWA\textsuperscript{\textdagger} & 87.0\tiny$\pm$0.5 & 77.2\tiny$\pm$0.6 & 69.5\tiny$\pm$0.4 & 50.1\tiny$\pm$0.7 & 44.0\tiny$\pm$0.2 & 65.6 & 293.1\\
& SelfReg + SWA & 86.5\tiny$\pm$0.3 & 77.5\tiny$\pm$0.0 & 69.4\tiny$\pm$0.2 & 51.0\tiny$\pm$0.4 & 44.6\tiny$\pm$0.1 & 65.8 & 327.1\\
& CORAL + SWA\textsuperscript{\textdagger} & 87.5\tiny$\pm$0.5 & 78.2\tiny$\pm$0.4 & 70.7\tiny$\pm$0.1 & 51.1\tiny$\pm$0.6 & 44.6\tiny$\pm$0.4 & 66.4 & 314.2\\
& FGMix + SWA & \textbf{88.4}\tiny$\pm$0.6 & \textbf{78.7}\tiny$\pm$0.6 & \textbf{71.3}\tiny$\pm$0.6 & \textbf{52.3}\tiny$\pm$0.9 & \textbf{45.1}\tiny$\pm$0.4 & \textbf{67.2} & 319.8\\
\bottomrule
\end{tabular}
\end{table}

The comparison results with baselines on the 5 datasets are presented in Table \ref{tbl:overall} (see \ref{appx:overall} for the disentangled results on each test domain). We observe that most of the DG algorithms outperform ERM in terms of average accuracy (except for AND-Mask). Our proposed FGMix achieves consistent performance gains over ERM for all 5 datasets: +1.3pp for \texttt{PACS}, +0.6pp for \texttt{VLCS}, +2.5pp for \texttt{OfficeHome}, +3.1pp for \texttt{TerraInc} and +1.2pp for \texttt{DomainNet}. We observe that FGMix performs exceptionally well on the more challenging datasets, where distribution shifts between source and target domains may be more pronounced, i.e., the gains are most pronounced for \texttt{OfficeHome} and \texttt{TerraInc}, whose test accuracies are relatively low compared to other datasets. Overall, our FGMix tops in 3 out of 5 datasets, achieving the best average accuracy with 0.4pp higher than the previous SOTA (i.e., CORAL), and 1.0pp higher than the strongest relevant baselines (i.e., Manifold Mixup, 
MetaMixup, Fishr, and L2A-OT).\

We further conduct experiments to combine FGMix with the flatness-aware optimizer SWA (Stochastic Weight Averaging) \citep{izmailov2018averaging,arpit2021ensemble}. SWA simply performs weight averaging, yielding a minimizer in the flatter central region of the found minimum. A recent study by \citet{arpit2021ensemble} found that SWA not only boosts the performance of DG algorithms but also ensures a better correlation between in-domain validation and out-domain test results, facilitating more reliable model selection. We combine SWA with FGMix and 3 other strong and representative DG algorithms (i.e., ERM, SelfReg, and CORAL) for comparison. Table \ref{tbl:overall} shows that FGMix + SWA achieves the best results for all 5 datasets. Notably, we observe that the performance gain is most pronounced for FGMix after combining with SWA, compared to SelfReg and CORAL (i.e., +1.8pp for SelfReg and CORAL, and +2.2pp for FGMix after combining with SWA). This shows that FGMix is more orthogonal to SWA than other DG methods, as its benefits persist even when combined with SWA.
To understand this, recall that FGMix is used to widen the loss surface for the optimizer to explore. While this increases the chance of covering the target area, it also enlarges the off-target area reachable to the optimizer. SWA with weight averaging along the training process helps mitigate the risk of the optimizer running into an undesirable region.\

In Table \ref{tbl:overall}, we also report the time cost (in ms) per training iteration averaged over 5000 iterations. The reported time is for datasets with 3 source domains and thus uses $k=3$ instances for mixup each time. We observe that the average per-iteration time cost for FGMix is only slightly higher than that of ERM. Among the competitive baselines (e.g., DomainBed SOTA), FGMix's time cost is on par with SelfReg and CORAL, and much lower than that of SagNet. Notably, when used with a flatness-aware optimizer such as SWA, the performance gain of FGMix becomes more pronounced, while the time cost remains the same for FGMix and other SOTA methods. Since the inference time cost of FGMix is the same as that of ERM (i.e., FGMix performs inference using only the trained base model), the minor additional costs incurred during training are considered acceptable given the evident performance gains.

%\vspace{-1mm}
\subsection{Ablation Study}
%\vspace{-2mm}
We test the efficacy of various components of FGMix by introducing 5 variants, each includes one additional component at a time: A is a simple interpolation baseline with random instance weights drawn from a Dirichlet distribution; B employs cosine similarity to compute the weights based on instances' latent codes (i.e., feature-based similarity);  C computes cosine similarity based on latent codes' gradients w.r.t. the classifier (i.e., gradient-based similarity); D further includes scaling and shifting for extrapolation by setting the scaling factor $\lambda$ as 3; E replaces the cosine similarity with a learnable compatibility score function $\mathcal{A}_{\omega}$ and learns both $\mathcal{A}_{\omega}$ and $\lambda$ towards flatter loss surface with the objective $\mathcal{L}_{flat}$. Further including the adversarial loss $\mathcal{L}_{adv}$ for prior matching results in our proposed FGMix. We also include a Variant F, which replaces the gradient-based inputs in FGMix with latent features, to examine feature-based conditioning for $\mathcal{A}_{\omega}$. \

\setlength{\tabcolsep}{4pt}
\begin{table}[t]
% \vspace{-5mm}
\caption{Ablation study on \texttt{PACS} and \texttt{TerraInc}. From A to FGMix we add one component at a time, where A is a simple interpolation baseline with random instance weights. For Variant F, we replace the gradient-based inputs in FGMix with latent features.}
%\vspace{-2mm}
\label{tbl:ablation}
\scriptsize
\centering
\begin{tabular}{c c c c c c c c}
%\begin{tabular}{|*{11}{c|}}
\toprule
\multirowcell{2}{\textbf{Variant}} & \multirowcell{2}{\scriptsize similarity-\\ \scriptsize based weights} & \multirowcell{2}{\scriptsize gradient-\\ \scriptsize based similarity} & \multirowcell{2}{\scriptsize scaling \\ \scriptsize \& shifting} & \multirowcell{2}{\scriptsize$\mathcal{L}_{flat}$} & \multirow{2}{*}{\scriptsize$\mathcal{L}_{adv}$} & \multirow{2}{*}{\texttt{PACS}} & \multirow{2}{*}{\texttt{TerraInc}} \\
% & \multirow{2}{*}{Avg.}\\
\\
\midrule
A (baseline) &&&&&& 84.7\tiny$\pm$0.7 & 47.5\tiny$\pm$1.1 \\
% & 66.1 \\
\midrule
B & \checkmark &&&&& 85.2\tiny$\pm$0.5 & 47.9\tiny$\pm$0.9 \\
% & 66.6\\
C & \checkmark & \checkmark &&&& 85.6\tiny$\pm$0.2 & 48.1\tiny$\pm$1.1 \\
% & 66.9 \\
D & \checkmark & \checkmark & \checkmark &&& 86.0\tiny$\pm$1.1 & 48.8\tiny$\pm$1.7 \\
% & 67.4\\
E & \checkmark & \checkmark & \checkmark & \checkmark && 86.7\tiny$\pm$1.7 & \textbf{49.2}\tiny$\pm$2.2 \\
F & \checkmark &  & \checkmark & \checkmark &  \checkmark & 86.2\tiny$\pm$1.5 & 48.9\tiny$\pm$1.7 \\ 
% & 68.0\\
\midrule
FGMix (ours) & \checkmark & \checkmark & \checkmark & \checkmark & \checkmark & \textbf{86.8}\tiny$\pm$1.2 & \textbf{49.2}\tiny$\pm$1.4 \\
% & 68.0\\
\bottomrule
\end{tabular}
%\vspace{-1mm}
\end{table}

Table \ref{tbl:ablation} presents the results (see \ref{appx:ablation} for results on each test domain). From A to FGMix, the incremental gains are +0.5pp, +0.4pp, +0.4pp, +0.7pp, +0.1pp for \texttt{PACS}, and +0.4pp, +0.2pp, +0.7pp, +0.4pp, +0.0pp for \texttt{TerraInc}. 
%This implies that the importance of various components follows the order: similarity-based weights $\approx$ learnable similarity ($\mathcal{L}_{flat}$) $>$ gradient-based similarity $\approx$ scaling \& shifting $>$ prior matching ($\mathcal{L}_{adv}$). 
Firstly, we see that enabling data extrapolation (i.e., scaling \& shifting) and learning towards a flatter loss surface (i.e., $\mathcal{L}_{flat}$) seem to be the two most important components, as both serve to avoid overfitting to the source domains and increase the chance of covering the target domain. The similarity-based weight computation that accounts for attention to other instances within the same combination also plays an important role, yielding evident improvements over the random weight baseline. Replacing feature-based similarity with gradient-based similarity also improves performance, as similarity in feature space does not characterize what the classifier deems invariant, and hence, weights based on feature similarity cannot identify instances that carry more invariant information. This trend is further confirmed by Variant F, which replaces gradient-based inputs in FGMix with latent features when learning $\mathcal{A}_{\omega}$. The fact that feature-based conditioning consistently underperforms FGMix highlights that gradients provide a more direct and informative signal for guiding flatness-aware mixup. Lastly, though matching the generated distribution to a prior (i.e., $\mathcal{L}_{adv}$) seem to have minor effects on the mean accuracy, it helps reduce the variance significantly from E to FGMix, demonstrating its role as a regularization measure to prevent over-extrapolation. In \ref{appx:sens}, we include a sensitivity test to evaluate the importance of $\mathcal{L}_{adv}$ by varying the weight $\alpha$.

\subsection{Qualitative Analysis}
\subsubsection{Mixup Distribution Visualization}

\begin{figure}[t]
\centering
\includegraphics[width=1.0\columnwidth]{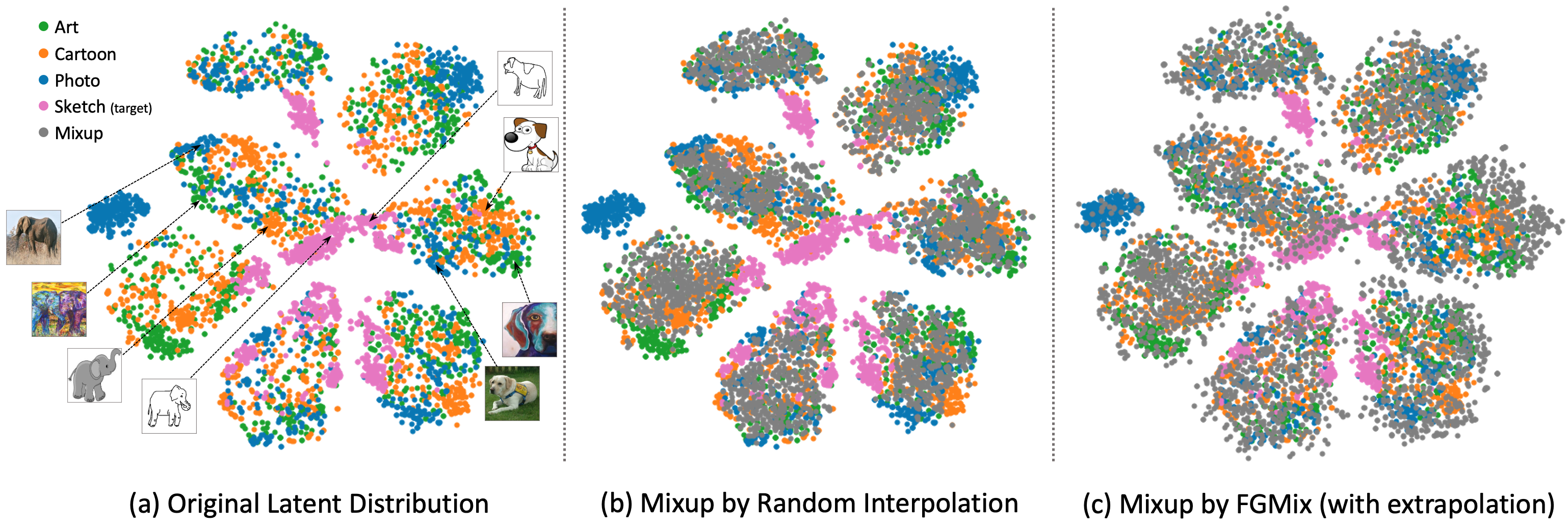}
\caption{t-SNE visualization of the latent distributions of \texttt{PACS} trained by using ``\texttt{Art}", ``\texttt{Cartoon}", ``\texttt{Photo}" as the source domains and ``\texttt{Sketch}" as the target domain. We plot the mixup distribution generated by random interpolation and by our FGMix which involves extrapolation.}
\label{fig:distribution}
\end{figure}

To understand how the generated distribution facilitates classifier learning, we visualize the latent codes of \texttt{PACS} in 2D using t-SNE. For this visualization, we train a model (i.e., feature extractor + classifier) with FGMix and obtain the latent codes from the feature extractor. To compare different mixup methods, we plot the mixup distribution generated by random interpolation (i.e., variant A) and by our FGMix, which involves extrapolation. Here, we use ``\texttt{Sketch}" as the target domain, as it is the most difficult domain when training with the other 3 domains.\

Figure \ref{fig:distribution} shows the distributions of the original latent codes (i.e., Figure \ref{fig:distribution}a), the mixup latent codes generated by random interpolation (i.e., Figure \ref{fig:distribution}b) and by FGMix (i.e., Figure \ref{fig:distribution}c), respectively. 
% Firstly, we see that by training with FGMix, 7 clusters corresponding to 7 class labels are clearly separated in the learned latent space, indicating that the class-discriminative features are well-captured by the feature extractor learned with FGMix. 
In Figure \ref{fig:distribution}a, we observe that the latent codes of the target domain ``\texttt{Sketch}" (i.e., the pink dots) are located near the decision boundary, especially for the ``dog" and ``elephant" classes, which are inseparable in the middle. In Figure \ref{fig:distribution}b, the mixup codes generated by random interpolation all lie within the convex hull formed by the source latent codes in the respective cluster, leaving the target region uncovered. Our FGMix (as shown in Figure \ref{fig:distribution}c), on the other hand, is able to generate codes outside of the convex hull with extrapolation, resulting in better coverage of the target region and hence better generalization performance of the classifier learned from the mixup latent codes. In this case, approximately 22.5\% of the generated mixup codes are extrapolated, and the remainder are interpolations. {One possible reason why $\mathcal{L}_{flat}$ can produce difficult samples near the decision boundary is that, as training progresses, increased prediction confidence flattens the local loss landscape, which may be associated with a smoother boundary and thus allows boundary-adjacent samples to remain robust to parameter perturbations.} In \ref{appx:vis}, we further include an analysis regarding the effect of mixup on the learned latent distributions. 

%\vspace{-1mm}
\subsubsection{Loss Landscape Visualization}
%\vspace{-2mm}
In this section, we verify that FGMix indeed generates flatter loss minima. We also observe how changes in the loss landscape affect the optimization process. The experiments are conducted on the \texttt{TerraInc} dataset.

% \setlength{\columnsep}{3.5mm}%
% \begin{wrapfigure}{r}{0.4\columnwidth}
% % \vspace{-5.2mm}
%   \begin{center}
%     \includegraphics[width=0.4\columnwidth]{figures/loss_curve_3}
%   \end{center}
%   % \vspace{-3mm}
%   \caption{We plot for 4 methods the square loss difference between the minimizer and its neighbourhoods at distance $\gamma=1, ..., 10$. The value is averaged over 50 random directions.}
% %  \hspace{-3mm}
%   % \vspace{-6mm}
%   \label{fig:loss_1d}
% \end{wrapfigure}

\begin{figure}[t]
\centering
\includegraphics[width=0.98\columnwidth]{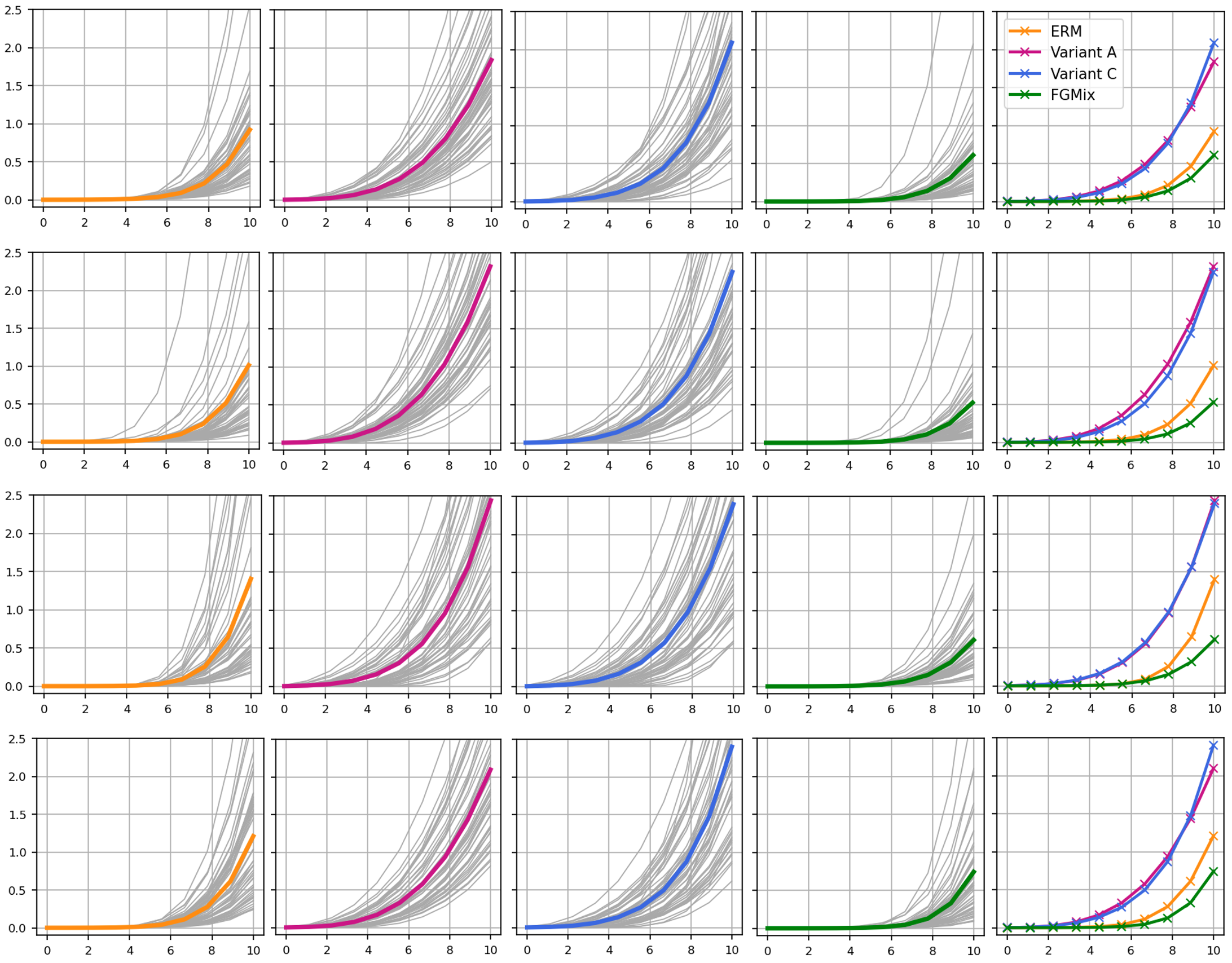}
\caption{Square loss difference between the classifier minimizer and its neighborhoods at distance $\gamma=1, ..., 10$ for ERM, variant A, variant C, and FGMix. Each row shows the results of using one of the 4 domains as the target domain (from the top down, the target domain is \texttt{L100}, \texttt{L38}, \texttt{L43}, and \texttt{L46}, respectively). We randomly sample 50 directions to compute the loss difference, indicated by the grey curves. The final results averaged over 50 random directions are compared in the last column.}
%\vspace{-1mm}
\label{fig:loss_1d}
\end{figure}

%\vspace{-3mm}
\paragraph{Flatness Analysis} We compare the flatness of FGMix minimizer with that of ERM, random mixup (i.e., variant A), and gradient-based similarity mixup (i.e., variant C) in Figure \ref{fig:loss_1d}. Following \citet{izmailov2018averaging,cha2021swad}, for each method, we plot the square loss difference between the minimizer $\phi^*$ and its neighborhoods $\phi'$ at distance $\gamma$, i.e., $\mathbb{E}_{\|\phi^*-\phi'\|=\gamma}[\mathcal{L}(\phi^*)-\mathcal{L}(\phi')]^2$, for $\gamma=1, ..., 10$. The expectation is approximated by averaging over 50 randomly sampled directions. Note that for ERM, the loss is computed based on the original training data, while for variant A, variant C, and FGMix, the loss is computed based on the mixture of original and mixup data.\

Figure \ref{fig:loss_1d} shows the separate plots for each of the 4 methods (including the results of the 50 sampled directions) and the combined plots. From top to bottom, each row shows the results obtained using \texttt{L100}, \texttt{L38}, \texttt{L43}, and \texttt{L46} as the target domain, respectively. First, by comparing variants A and C with ERM, we see that simply adding mixup data yields a sharper loss minimum. Fortunately, our FGMix with the flatness-aware objective effectively reduces loss sharpness by adjusting the weight generation policy, yielding a minimum that is even flatter than that of ERM. 
\begin{figure}[t]
\centering
\includegraphics[width=\columnwidth]{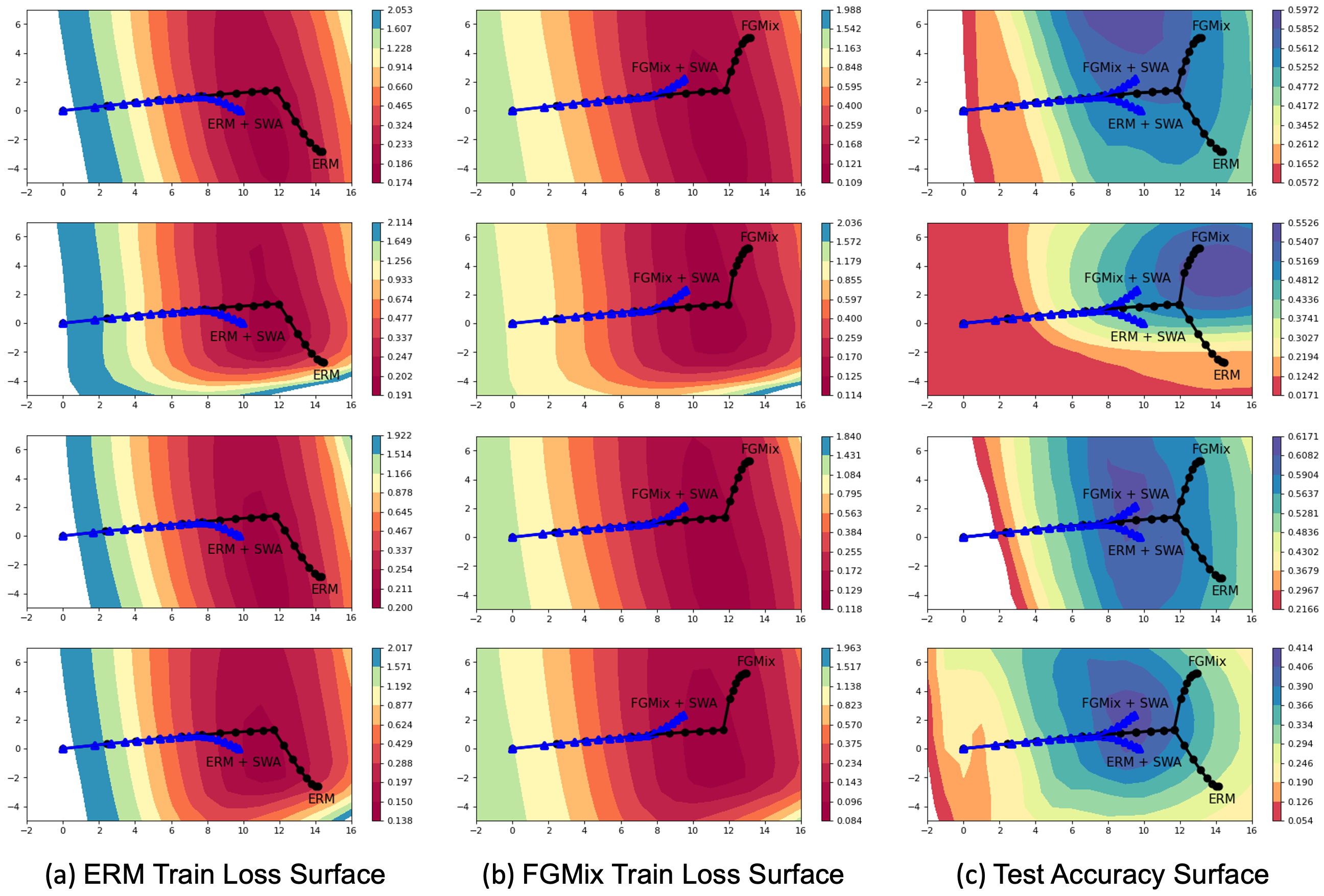}
\caption{2D loss and accuracy surfaces generated using 3 models at the initialization, the end of ERM + SWA, and the end of FGMix + SWA, respectively. Each row shows the results obtained by using one of the 4 domains as the target domain (from top to bottom, the target domains are \texttt{L100}, \texttt{L38}, \texttt{L43}, and \texttt{L46}, respectively). Note that only the 3 models used to generate the plots are on the 2D plane (i.e., the loss/accuracy values reported for them are correct). Also note that for FGMix, since the mixup data is only added for training from the 3,000th iteration onwards, the loss values indicated for the optimization path before the 3,000th iteration in (b) do not reflect the actual loss during training. For direct comparison of the flatness, we use the same scale interval for (a) and (b).}
\label{fig:loss_2d}
\end{figure}
% \captionsetup[figure]{labelfont={color=black}}

\paragraph{Loss Surface \& Optimization Path} To observe how the generated data affects the optimization process, in Figure \ref{fig:loss_2d}, we plot the 2D loss surface of the original training data (i.e., ERM) as well as the one after adding the generated mixup data\footnote{The mixup data used here is generated by the well-trained $\mathcal{A}_{\omega}$ and scaling factor $\lambda$ at the last iteration.} (i.e., FGMix). We also plot the accuracy surface on the test data to visualize distribution shifts relative to the training domains. For the optimization paths, we consider both the original paths of ERM and FGMix and the ones with SWA, as weight averaging helps stabilize the optimization process and converge to a better minimum. To generate the 2D loss surface, we follow \citet{izmailov2018averaging} by choosing 3 models to compute the orthonormal basis of the 2D plane containing the 3 models (computation details are included in \ref{appx:loss_surface_const}). 
For a better span of the 2D perspective, the 3 models chosen are 1) the common initialization, 2) the final checkpoint of ERM + SWA, and 3) the final checkpoint of FGMix + SWA. 

Figure \ref{fig:loss_2d}a and \ref{fig:loss_2d}b show the train loss surfaces and optimization paths for ERM and FGMix, respectively, and Figure \ref{fig:loss_2d}c shows the test accuracy surface with paths of both ERM and FGMix on it. From top to bottom, each row presents the results obtained using \texttt{L100}, \texttt{L38}, \texttt{L43}, and \texttt{L46} as the target domain, respectively. First, note that, since the generated data is included in model training only from the 3,000th iteration onwards, ERM and FGMix follow the same path before the 3,000th step. Comparing Figure \ref{fig:loss_2d}a and \ref{fig:loss_2d}c, we see that there is a clear shift in optimum between the ERM train loss and the test accuracy, which results in the ERM optimization path converging to a point far from the test optimum. While for FGMix (as shown in Figure \ref{fig:loss_2d}b), the loss surface is flatter as compared to that of ERM after adding the mixup data, yielding a broader minimum that better covers the test optimum. As a result, the FGMix optimization path converges to a point closer to the test optimum; in Figure \ref{fig:loss_2d}c, FGMix + SWA terminates in a region with higher test accuracy than ERM + SWA. The result is consistent across all 4 experiments with different target domains.

%\vspace{-3mm}
\section{Conclusion and Discussion}
%\vspace{-3mm}
In this work, we explore a mixup method with data extrapolation. We propose a weight generation policy, FGMix, that computes instance weights from gradients and learns toward flatter minima. We also employ an adversarial loss for regularization. Extensive experiments on the DomainBed benchmark demonstrate FGMix's effectiveness. Perhaps the major limitation of FGMix is the large variance caused by data extrapolation, as we lack strategies to guide extrapolation toward the expected region. In the future, we will consider exploiting domain-specific information and relationships among the source domains to design a more effective extrapolation strategy.\

\section*{Acknowledgement}
Sinno J. Pan thanks the support from the JC STEM Lab of Machine Learning and Symbolic Reasoning funded by The Hong Kong Jockey Club Charities Trust.

%\clearpage

\bibliographystyle{elsarticle-num-names}
\bibliography{references}

@inproceedings{berthelot2018understanding,
  title={Understanding and Improving Interpolation in Autoencoders via an Adversarial Regularizer},
  author={Berthelot, David and Raffel, Colin and Roy, Aurko and Goodfellow, Ian},
  booktitle={International Conference on Learning Representations},
  year={2018}
}

@inproceedings{zhang2018mixup,
  title={mixup: Beyond Empirical Risk Minimization},
  author={Zhang, Hongyi and Cisse, Moustapha and Dauphin, Yann N and Lopez-Paz, David},
  booktitle={International Conference on Learning Representations},
  year={2018}
}

@inproceedings{zhou2020domain,
  title={Domain Generalization with MixStyle},
  author={Zhou, Kaiyang and Yang, Yongxin and Qiao, Yu and Xiang, Tao},
  booktitle={International Conference on Learning Representations},
  year={2020}
}

@inproceedings{wang2020heterogeneous,
  title={Heterogeneous domain generalization via domain mixup},
  author={Wang, Yufei and Li, Haoliang and Kot, Alex C},
  booktitle={ICASSP 2020-2020 IEEE International Conference on Acoustics, Speech and Signal Processing (ICASSP)},
  pages={3622--3626},
  year={2020},
  organization={IEEE}
}

@article{mai2021metamixup,
  title={MetaMixUp: Learning Adaptive Interpolation Policy of MixUp With Metalearning},
  author={Mai, Zhijun and Hu, Guosheng and Chen, Dexiong and Shen, Fumin and Shen, Heng Tao},
  journal={IEEE Transactions on Neural Networks and Learning Systems},
  year={2021},
  publisher={IEEE}
}

@inproceedings{verma2019manifold,
  title={Manifold mixup: Better representations by interpolating hidden states},
  author={Verma, Vikas and Lamb, Alex and Beckham, Christopher and Najafi, Amir and Mitliagkas, Ioannis and Lopez-Paz, David and Bengio, Yoshua},
  booktitle={International Conference on Machine Learning},
  pages={6438--6447},
  year={2019},
  organization={PMLR}
}

@inproceedings{parascandolo2020learning,
  title={Learning explanations that are hard to vary},
  author={Parascandolo, Giambattista and Neitz, Alexander and ORVIETO, ANTONIO and Gresele, Luigi and Sch{\"o}lkopf, Bernhard},
  booktitle={International Conference on Learning Representations},
  year={2020}
}

@article{teterwak2023erm++,
  title={ERM++: An Improved Baseline for Domain Generalization},
  author={Teterwak, Piotr and Saito, Kuniaki and Tsiligkaridis, Theodoros and Saenko, Kate and Plummer, Bryan A},
  journal={arXiv preprint arXiv:2304.01973},
  year={2023}
}

@article{shi2021gradient,
  title={Gradient Matching for Domain Generalization},
  author={Shi, Yuge and Seely, Jeffrey and Torr, Philip HS and Siddharth, N and Hannun, Awni and Usunier, Nicolas and Synnaeve, Gabriel},
  journal={arXiv preprint arXiv:2104.09937},
  year={2021}
}

@inproceedings{sung2018learning,
  title={Learning to compare: Relation network for few-shot learning},
  author={Sung, Flood and Yang, Yongxin and Zhang, Li and Xiang, Tao and Torr, Philip HS and Hospedales, Timothy M},
  booktitle={Proceedings of the IEEE conference on computer vision and pattern recognition},
  pages={1199--1208},
  year={2018}
}

@inproceedings{he2017neural,
  title={Neural collaborative filtering},
  author={He, Xiangnan and Liao, Lizi and Zhang, Hanwang and Nie, Liqiang and Hu, Xia and Chua, Tat-Seng},
  booktitle={Proceedings of the 26th international conference on world wide web},
  pages={173--182},
  year={2017}
}

@inproceedings{li2018learning,
  title={Learning to generalize: Meta-learning for domain generalization},
  author={Li, Da and Yang, Yongxin and Song, Yi-Zhe and Hospedales, Timothy M},
  booktitle={Thirty-Second AAAI Conference on Artificial Intelligence},
  year={2018}
}

@article{makhzani2015adversarial,
  title={Adversarial autoencoders},
  author={Makhzani, Alireza and Shlens, Jonathon and Jaitly, Navdeep and Goodfellow, Ian and Frey, Brendan},
  journal={arXiv preprint arXiv:1511.05644},
  year={2015}
}

@article{blanchard2011generalizing,
  title={Generalizing from several related classification tasks to a new unlabeled sample},
  author={Blanchard, Gilles and Lee, Gyemin and Scott, Clayton},
  journal={Advances in neural information processing systems},
  volume={24},
  year={2011}
}

@inproceedings{muandet2013domain,
  title={Domain generalization via invariant feature representation},
  author={Muandet, Krikamol and Balduzzi, David and Sch{\"o}lkopf, Bernhard},
  booktitle={International Conference on Machine Learning},
  pages={10--18},
  year={2013},
  organization={PMLR}
}

@inproceedings{li2019episodic,
  title={Episodic training for domain generalization},
  author={Li, Da and Zhang, Jianshu and Yang, Yongxin and Liu, Cong and Song, Yi-Zhe and Hospedales, Timothy M},
  booktitle={Proceedings of the IEEE/CVF International Conference on Computer Vision},
  pages={1446--1455},
  year={2019}
}

@inproceedings{li2017deeper,
  title={Deeper, broader and artier domain generalization},
  author={Li, Da and Yang, Yongxin and Song, Yi-Zhe and Hospedales, Timothy M},
  booktitle={Proceedings of the IEEE international conference on computer vision},
  pages={5542--5550},
  year={2017}
}

@article{gulrajani2020search,
  title={In search of lost domain generalization},
  author={Gulrajani, Ishaan and Lopez-Paz, David},
  journal={arXiv preprint arXiv:2007.01434},
  year={2020}
}

@inproceedings{cha2021swad,
  title={SWAD: Domain Generalization by Seeking Flat Minima},
  author={Cha, Junbum and Chun, Sanghyuk and Lee, Kyungjae and Cho, Han-Cheol and Park, Seunghyun and Lee, Yunsung and Park, Sungrae},
  booktitle={Advances in Neural Information Processing Systems},
  year={2021}
}

@article{arpit2021ensemble,
  title={Ensemble of Averages: Improving Model Selection and Boosting Performance in Domain Generalization},
  author={Arpit, Devansh and Wang, Huan and Zhou, Yingbo and Xiong, Caiming},
  journal={arXiv preprint arXiv:2110.10832},
  year={2021}
}

@inproceedings{keskar2017large,
  title={On large-batch training for deep learning: Generalization gap and sharp minima},
  author={Keskar, Nitish Shirish and Nocedal, Jorge and Tang, Ping Tak Peter and Mudigere, Dheevatsa and Smelyanskiy, Mikhail},
  booktitle={5th International Conference on Learning Representations, ICLR 2017},
  year={2017}
}

@article{hochreiter1997flat,
  title={FLAT MINIMA},
  author={Hochreiter, Sepp and urgen Schmidhuber, J and Elvezia, Corso},
  journal={Neural Computation},
  volume={9},
  number={1},
  pages={1--42},
  year={1997}
}

@inproceedings{izmailov2018averaging,
  title={Averaging weights leads to wider optima and better generalization},
  author={Izmailov, P and Wilson, AG and Podoprikhin, D and Vetrov, D and Garipov, T},
  booktitle={34th Conference on Uncertainty in Artificial Intelligence 2018, UAI 2018},
  pages={876--885},
  year={2018}
}

@inproceedings{foret2020sharpness,
  title={Sharpness-aware Minimization for Efficiently Improving Generalization},
  author={Foret, Pierre and Kleiner, Ariel and Mobahi, Hossein and Neyshabur, Behnam},
  booktitle={International Conference on Learning Representations},
  year={2020}
}

@inproceedings{petzka2021relative,
  title={Relative Flatness and Generalization},
  author={Petzka, Henning and Kamp, Michael and Adilova, Linara and Sminchisescu, Cristian and Boley, Mario},
  booktitle={Advances in Neural Information Processing Systems},
  year={2021}
}

@inproceedings{li2018domain,
  title={Domain generalization with adversarial feature learning},
  author={Li, Haoliang and Pan, Sinno Jialin and Wang, Shiqi and Kot, Alex C},
  booktitle={Proceedings of the IEEE conference on computer vision and pattern recognition},
  pages={5400--5409},
  year={2018}
}

@inproceedings{zhou2020learning,
  title={Learning to generate novel domains for domain generalization},
  author={Zhou, Kaiyang and Yang, Yongxin and Hospedales, Timothy and Xiang, Tao},
  booktitle={European conference on computer vision},
  pages={561--578},
  year={2020},
  organization={Springer}
}

@inproceedings{zhang2022towards,
  title={Towards principled disentanglement for domain generalization},
  author={Zhang, Hanlin and Zhang, Yi-Fan and Liu, Weiyang and Weller, Adrian and Sch{\"o}lkopf, Bernhard and Xing, Eric P},
  booktitle={Proceedings of the IEEE/CVF Conference on Computer Vision and Pattern Recognition},
  pages={8024--8034},
  year={2022}
}

@inproceedings{balaji2018metareg,
  title={MetaReg: towards domain generalization using meta-regularization},
  author={Balaji, Yogesh and Sankaranarayanan, Swami and Chellappa, Rama},
  booktitle={Proceedings of the 32nd International Conference on Neural Information Processing Systems},
  pages={1006--1016},
  year={2018}
}

@inproceedings{tang2021crossnorm,
  title={Crossnorm and selfnorm for generalization under distribution shifts},
  author={Tang, Zhiqiang and Gao, Yunhe and Zhu, Yi and Zhang, Zhi and Li, Mu and Metaxas, Dimitris N},
  booktitle={Proceedings of the IEEE/CVF International Conference on Computer Vision},
  pages={52--61},
  year={2021}
}

@inproceedings{chapelle2000vicinal,
  title={Vicinal Risk Minimization},
  author={Chapelle, Olivier and Weston, Jason and Bottou, L{\'e}on and Vapnik, Vladimir},
  booktitle={NIPS},
  year={2000}
}

@inproceedings{yun2019cutmix,
  title={Cutmix: Regularization strategy to train strong classifiers with localizable features},
  author={Yun, Sangdoo and Han, Dongyoon and Oh, Seong Joon and Chun, Sanghyuk and Choe, Junsuk and Yoo, Youngjoon},
  booktitle={Proceedings of the IEEE/CVF international conference on computer vision},
  pages={6023--6032},
  year={2019}
}

@inproceedings{chou2020remix,
  title={Remix: rebalanced mixup},
  author={Chou, Hsin-Ping and Chang, Shih-Chieh and Pan, Jia-Yu and Wei, Wei and Juan, Da-Cheng},
  booktitle={European Conference on Computer Vision},
  pages={95--110},
  year={2020},
  organization={Springer}
}

@inproceedings{guo2019mixup,
  title={Mixup as locally linear out-of-manifold regularization},
  author={Guo, Hongyu and Mao, Yongyi and Zhang, Richong},
  booktitle={Proceedings of the AAAI Conference on Artificial Intelligence},
  volume={33},
  number={01},
  pages={3714--3722},
  year={2019}
}

@inproceedings{riemer2018learning,
  title={Learning to Learn without Forgetting by Maximizing Transfer and Minimizing Interference},
  author={Riemer, Matthew and Cases, Ignacio and Ajemian, Robert and Liu, Miao and Rish, Irina and Tu, Yuhai and Tesauro, Gerald},
  booktitle={International Conference on Learning Representations},
  year={2018}
}

@article{yu2020gradient,
  title={Gradient surgery for multi-task learning},
  author={Yu, Tianhe and Kumar, Saurabh and Gupta, Abhishek and Levine, Sergey and Hausman, Karol and Finn, Chelsea},
  journal={Advances in Neural Information Processing Systems},
  volume={33},
  pages={5824--5836},
  year={2020}
}

@inproceedings{mansilla2021domain,
  title={Domain generalization via gradient surgery},
  author={Mansilla, Lucas and Echeveste, Rodrigo and Milone, Diego H and Ferrante, Enzo},
  booktitle={Proceedings of the IEEE/CVF International Conference on Computer Vision},
  pages={6630--6638},
  year={2021}
}

@article{shahtalebi2021sand,
  title={Sand-mask: An enhanced gradient masking strategy for the discovery of invariances in domain generalization},
  author={Shahtalebi, Soroosh and Gagnon-Audet, Jean-Christophe and Laleh, Touraj and Faramarzi, Mojtaba and Ahuja, Kartik and Rish, Irina},
  journal={arXiv preprint arXiv:2106.02266},
  year={2021}
}

@article{rame2021fishr,
  title={Fishr: Invariant gradient variances for out-of-distribution generalization},
  author={Rame, Alexandre and Dancette, Corentin and Cord, Matthieu},
  journal={arXiv preprint arXiv:2109.02934},
  year={2021}
}

@inproceedings{nam2021reducing,
  title={Reducing domain gap by reducing style bias},
  author={Nam, Hyeonseob and Lee, HyunJae and Park, Jongchan and Yoon, Wonjun and Yoo, Donggeun},
  booktitle={Proceedings of the IEEE/CVF Conference on Computer Vision and Pattern Recognition},
  pages={8690--8699},
  year={2021}
}

@book{vapnick1998statistical,
  title={Statistical learning theory},
  author={Vapnick, Vladimir N},
  year={1998},
  publisher={Wiley, New York}
}

@inproceedings{kim2021selfreg,
  title={Selfreg: Self-supervised contrastive regularization for domain generalization},
  author={Kim, Daehee and Yoo, Youngjun and Park, Seunghyun and Kim, Jinkyu and Lee, Jaekoo},
  booktitle={Proceedings of the IEEE/CVF International Conference on Computer Vision},
  pages={9619--9628},
  year={2021}
}

@article{mackay1992practical,
  title={A practical Bayesian framework for backpropagation networks},
  author={MacKay, David JC},
  journal={Neural Computation},
  volume={4},
  number={3},
  pages={448--472},
  year={1992},
  publisher={MIT Press Cambridge, MA, USA}
}

@inproceedings{chaudhari2019entropy,
  title={Entropy-SGD: Biasing gradient descent into wide valleys (International Conference on Learning Representations, ICLR 2017)},
  author={Chaudhari, Pratik and Choromanska, Anna and Soatto, Stefano and LeCun, Yann and Baldassi, Carlo and Borgs, Christian and Chayes, Jennifer and Sagun, Levent and Zecchina, Riccardo},
  booktitle={5th International Conference on Learning Representations, ICLR 2017},
  year={2019}
}

@article{guo2022stochastic,
  title={Stochastic Weight Averaging Revisited},
  author={Guo, Hao and Jin, Jiyong and Liu, Bin},
  journal={arXiv preprint arXiv:2201.00519},
  year={2022}
}

@inproceedings{fang2013unbiased,
  title={Unbiased metric learning: On the utilization of multiple datasets and web images for softening bias},
  author={Fang, Chen and Xu, Ye and Rockmore, Daniel N},
  booktitle={Proceedings of the IEEE International Conference on Computer Vision},
  pages={1657--1664},
  year={2013}
}

@inproceedings{venkateswara2017deep,
  title={Deep hashing network for unsupervised domain adaptation},
  author={Venkateswara, Hemanth and Eusebio, Jose and Chakraborty, Shayok and Panchanathan, Sethuraman},
  booktitle={Proceedings of the IEEE conference on computer vision and pattern recognition},
  pages={5018--5027},
  year={2017}
}

@inproceedings{beery2018recognition,
  title={Recognition in terra incognita},
  author={Beery, Sara and Van Horn, Grant and Perona, Pietro},
  booktitle={Proceedings of the European conference on computer vision (ECCV)},
  pages={456--473},
  year={2018}
}

@inproceedings{peng2019moment,
  title={Moment matching for multi-source domain adaptation},
  author={Peng, Xingchao and Bai, Qinxun and Xia, Xide and Huang, Zijun and Saenko, Kate and Wang, Bo},
  booktitle={Proceedings of the IEEE/CVF international conference on computer vision},
  pages={1406--1415},
  year={2019}
}

@inproceedings{sun2016deep,
  title={Deep coral: Correlation alignment for deep domain adaptation},
  author={Sun, Baochen and Saenko, Kate},
  booktitle={European conference on computer vision},
  pages={443--450},
  year={2016},
  organization={Springer}
}

@inproceedings{he2016deep,
  title={Deep residual learning for image recognition},
  author={He, Kaiming and Zhang, Xiangyu and Ren, Shaoqing and Sun, Jian},
  booktitle={Proceedings of the IEEE conference on computer vision and pattern recognition},
  pages={770--778},
  year={2016}
}

@article{giannone2022just,
  title={Just Mix Once: Worst-group Generalization by Group Interpolation},
  author={Giannone, Giorgio and Havrylov, Serhii and Massiah, Jordan and Yilmaz, Emine and Jiao, Yunlong},
  journal={arXiv preprint arXiv:2210.12195},
  year={2022}
}

@article{hwang2022selecmix,
  title={Selecmix: Debiased learning by contradicting-pair sampling},
  author={Hwang, Inwoo and Lee, Sangjun and Kwak, Yunhyeok and Oh, Seong Joon and Teney, Damien and Kim, Jin-Hwa and Zhang, Byoung-Tak},
  journal={Advances in Neural Information Processing Systems},
  volume={35},
  pages={14345--14357},
  year={2022}
}

@inproceedings{yao2022improving,
  title={Improving out-of-distribution robustness via selective augmentation},
  author={Yao, Huaxiu and Wang, Yu and Li, Sai and Zhang, Linjun and Liang, Weixin and Zou, James and Finn, Chelsea},
  booktitle={International Conference on Machine Learning},
  pages={25407--25437},
  year={2022},
  organization={PMLR}
}

@article{goodfellow2014generative,
  title={Generative adversarial nets},
  author={Goodfellow, Ian and Pouget-Abadie, Jean and Mirza, Mehdi and Xu, Bing and Warde-Farley, David and Ozair, Sherjil and Courville, Aaron and Bengio, Yoshua},
  journal={Advances in neural information processing systems},
  volume={27},
  year={2014}
}

@article{eshratifar2018gradient,
  title={Gradient agreement as an optimization objective for meta-learning},
  author={Eshratifar, Amir Erfan and Eigen, David and Pedram, Massoud},
  journal={arXiv preprint arXiv:1810.08178},
  year={2018}
}

@article{du2018adapting,
  title={Adapting auxiliary losses using gradient similarity},
  author={Du, Yunshu and Czarnecki, Wojciech M and Jayakumar, Siddhant M and Farajtabar, Mehrdad and Pascanu, Razvan and Lakshminarayanan, Balaji},
  journal={arXiv preprint arXiv:1812.02224},
  year={2018}
}

\newpage
\appendix

%\newpage

\section{On the Connection with other Gradient Methods}
\label{appx:toy}
Like our proposed method for quantifying the amount of invariant information based on the sum of gradient similarities, a few recent DG works on gradient surgery also place greater emphasis on examples whose gradients align more closely with those of other examples, aiming to capture more invariant features in the model \cite{mansilla2021domain,parascandolo2020learning}. In \cite{mansilla2021domain}, the authors empirically show that the gradient cosine similarity is higher for inter-domain pairwise example comparisons as opposed to the intra-domain comparisons. They subsequently propose zeroing out or randomizing the gradient component when there is a conflicting sign across gradients between different domains. AND-Mask \cite{parascandolo2020learning} performs a similar procedure, except that it loosens the criteria by zeroing out gradient components only when the percentage of disagreeing signs is greater than a threshold. These methods implicitly place more emphasis on more aligned gradients, as the retained gradient components are largely contributed by them, which constitute the majority of the population.\

\begin{figure}[ht]
\centerline{\includegraphics[width=0.5\linewidth]{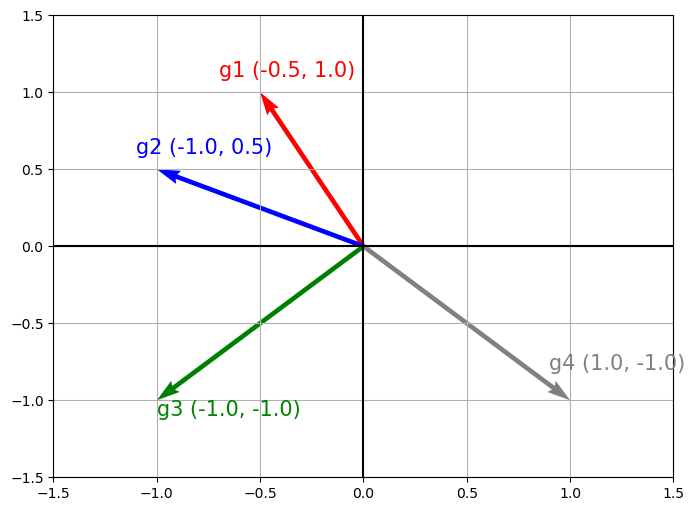}}
\caption{Gradients of 4 examples from different domains.}
\label{fig:toy}
\end{figure}

In the following, we present a toy example to illustrate the analogy between our proposed FGMix and AND-Mask \cite{parascandolo2020learning}, yielding a similar gradient update direction for learning the model. Figure \ref{fig:toy} shows 4 gradients of 4 examples belonging to different domains, i.e., $\mathbf{g}_1=[-0.5, 1], \mathbf{g}_2=[-1, 0.5], \mathbf{g}_3=[-1, -1], \mathbf{g}_4=[1, -1]$. Standard empirical risk minimization (ERM) updates the model using the overall gradient $\mathbf{g}_{ERM}=\sum_i \mathbf{g}_i = [-1.5, -0.5]$. AND-Mask eliminates the gradient component when the percentage of disagreeing signs exceeds a pre-defined threshold. Suppose we set the threshold at $30\%$. The first gradient component has $25\%$ disagreeing signs, i.e., $\textrm{sign}(\mathbf{g}_1^1)=\textrm{sign}(\mathbf{g}_2^1)=\textrm{sign}(\mathbf{g}_3^1)\neq\textrm{sign}(\mathbf{g}_4^1)$. The second gradient component has $50\%$ disagreeing signs, i.e., $\textrm{sign}(\mathbf{g}_1^2)=\textrm{sign}(\mathbf{g}_2^2)\neq\textrm{sign}(\mathbf{g}_3^2)=\textrm{sign}(\mathbf{g}_4^2)$. Hence, the first component is retained, while the second component is replaced with zero, resulting in the overall gradient $\mathbf{g}_{AND}=[-1.5, 0.0]$. For our FGMix, we compute the weight associated with $\mathbf{g}_i$ based on the sum of cosine similarities $s_i=\sum_{j\neq i}\cos{\psi_{i,j}}$ followed by softmax normalization. The computation is shown in Table \ref{tbl:toy}. Based on the weight $a_i$ computed, the aggregated gradient of FGMix is $\mathbf{g}_{FGMix}=[-0.793, 0.023]$.\

\begin{table}[ht]
\footnotesize
\centering
\caption{FGMix weight computation based on the sum of gradient cosine similarities.}
%\small
\begin{tabular}{c c c c c c c c}
\toprule
 &  & $\cos{\psi_{i,1}}$ & $\cos{\psi_{i,2}}$ & $\cos{\psi_{i,3}}$ & $\cos{\psi_{i,4}}$ & $s_i=\sum_{j\neq i}\cos{\psi_{i,j}}$ & $a_i=\frac{\exp(s_i)}{\sum_{i'=1}^k\exp(s_{i'})}$\\
\midrule
$\mathbf{g}_1$ & [-0.5, 1] & - & 0.8 & -0.316 & -0.949 & -0.465 & 0.21218712 \\
$\mathbf{g}_2$ & [-1, 0.5] & 0.8 & - & 0.316 & -0.949 & 0.168 & 0.39938428 \\
$\mathbf{g}_3$ & [-1, -1] & -0.316 & 0.316 & - & 0.0 & 0.0 & 0.33777486 \\
$\mathbf{g}_4$ & [1, -1] & -0.949 & -0.949 & 0.0 & - & -1.897 & 0.05065374 \\
\bottomrule
\label{tbl:toy}
\end{tabular}
\end{table}

To summarize, the gradients obtained using the 3 methods are:
\begin{equation*}
	\begin{aligned}
	&\mathbf{g}_{ERM}=[-1.5, -0.5], \quad \mathbf{g}_{AND}=[-1.5, 0.0], \quad \mathbf{g}_{FGMix}=[-0.793, 0.023].
	\end{aligned}
\end{equation*}
By comparing $\mathbf{g}_{FGMix}$ with $\mathbf{g}_{ERM}$ and $\mathbf{g}_{AND}$, we find that the cosine similarity between $\mathbf{g}_{FGMix}$ and $\mathbf{g}_{AND}$ is higher than that between $\mathbf{g}_{FGMix}$ and $\mathbf{g}_{ERM}$ , i.e., $\cos{\psi_{FGMix,AND}}=0.9996, \ \cos{\psi_{FGMix,ERM}}=0.9389$. This suggests that FGMix produces gradient update with direction similar to that of AND-Mask, which share the same goal of capturing invariant information for better generalizing across domains.\

Note that although $\mathbf{g}_{FGMix}$ and $\mathbf{g}_{AND}$ exhibit high similarity in this toy example, our actual FGMix algorithm functions differently and achieves different effects, as it employs a learnable similarity function $\mathcal{A}_{\omega}$ instead of the pre-defined cosine similarity function used in this example. We use cosine similarity to illustrate our motivation for simplicity; the same idea applies to other similarity metrics.

\newpage
\section{On the Connection between $\mathcal{A}_{\omega}(\cdot)$ and Cosine Similarity}
\label{appx:corr}
As discussed in the main text, cosine similarity is introduced primarily as a motivating example to provide intuition. Since $\mathcal{A}_{\omega}(\cdot)$ is a learned function optimized toward the designed objectives, it is not guaranteed to behave exactly like cosine similarity or any other conventional geometric similarity metric. Instead, the role of $\mathcal{A}_{\omega}$ is to assign higher scores to pairs of gradients whose joint effect promotes flatter optimization landscapes under mild prior-matching regularization. \

The behavior of $\mathcal{A}_{\omega}$ is mainly shaped by the flatness-aware loss in Eq. (6), which favors mixup samples whose induced gradients remain stable under parameter perturbations. Under such an objective, gradient pairs with lower cosine similarity or stronger misalignment are relatively less favored, as they are more likely to induce higher gradient variance, sharper curvature, and greater loss sensitivity—behaviors penalized by the flatness objective. Consequently, although $\mathcal{A}_{\omega}$ is unconstrained in form, the optimization pressure biases it toward assigning higher scores to more aligned gradient pairs. This leads to similarity-like behavior \textit{in expectation}, while still allowing flexibility beyond conventional similarity measures.\

\begin{figure}
\centering
\includegraphics[width=1\columnwidth]{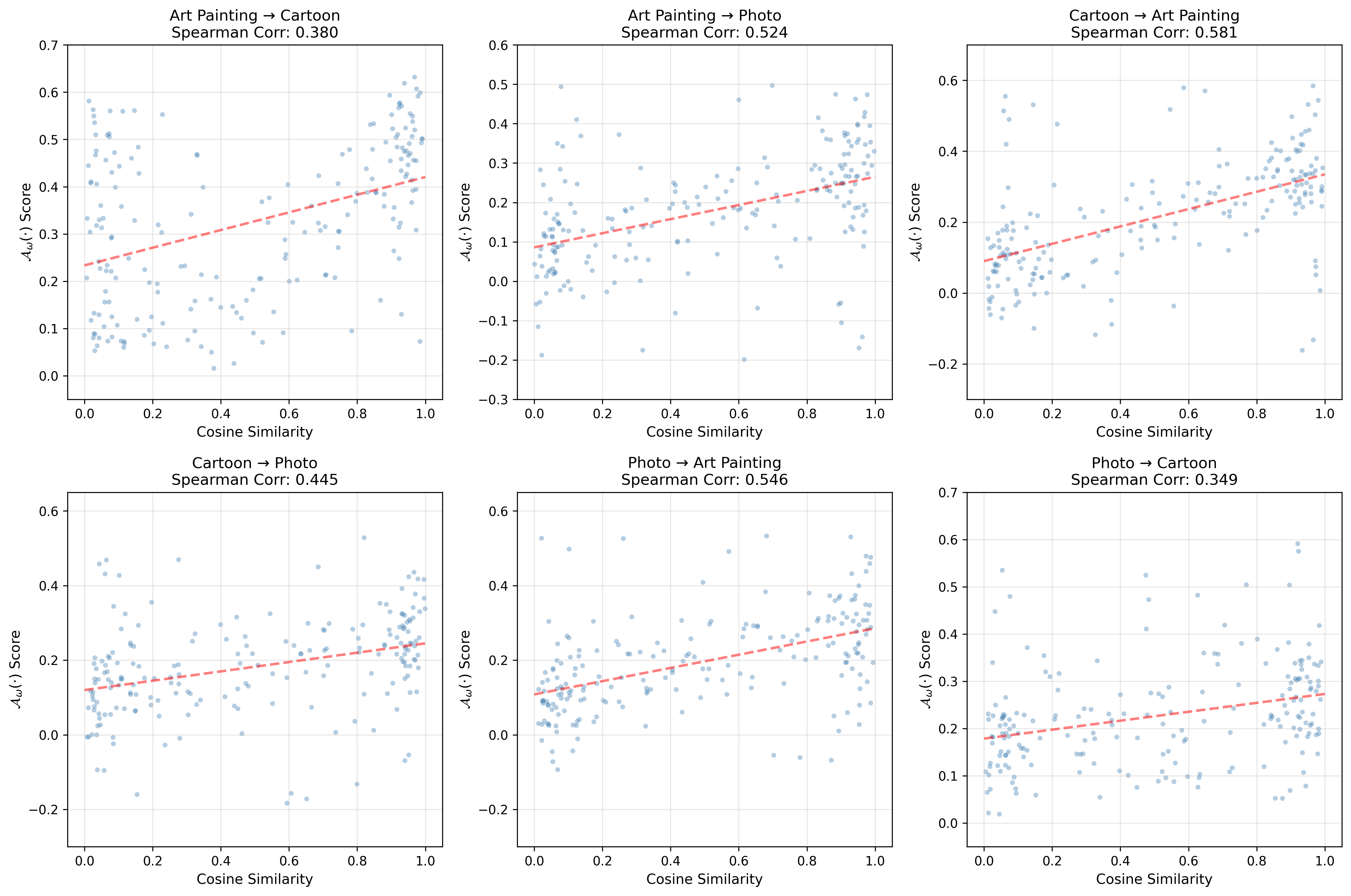}
\caption{Scatter plots of learned $\mathcal{A}_{\omega}(\cdot)$ scores versus cosine similarity for 200 randomly sampled cross-domain gradient pairs across six cross-domain settings (excluding the target domain) on the \texttt{PACS} dataset.}
\label{fig:corr_scatter_plots}
\end{figure}

To empirically examine this behavior, we analyze the relationship between the learned $\mathcal{A}_{\omega}(\cdot)$ scores and cosine similarity using 200 randomly sampled cross-domain gradient pairs across six cross-domain settings (excluding the target domain) on the PACS dataset. Figure \ref{fig:corr_scatter_plots} presents scatter plots of $\mathcal{A}_{\omega}$ versus cosine similarity for each domain pair. We observe consistent positive Spearman correlations, ranging from approximately 0.35 to 0.58 across all six settings, indicating that gradient pairs with higher cosine similarity tend to receive higher $\mathcal{A}_{\omega}$ scores. At the same time, some gradient pairs with low cosine similarity can still obtain high $\mathcal{A}_{\omega}$ scores, showing that such cases are not strictly excluded. This suggests that $\mathcal{A}_{\omega}$ captures additional factors beyond pure geometric similarity, consistent with its role as a learned compatibility measure rather than a fixed similarity metric.\

\newpage
\section{Additional Experimental Results}
\subsection{Full Results for Overall Comparison}
\label{appx:overall}

\setlength{\tabcolsep}{3pt}
\begin{table}[!h]
\caption{Overall comparison of selected algorithms on \texttt{PACS}.}
\label{tbl:overall_pacs}
\scriptsize
\centering
\begin{tabular}{c l | c c c c | c}
\toprule
& \textbf{Algorithm} & \texttt{A} & \texttt{C} & \texttt{P} & \texttt{S} & Avg. \\
\midrule
& ERM \citep{vapnick1998statistical} & 84.7\tiny$\pm$0.4 & 80.8\tiny$\pm$0.6 & 97.2\tiny$\pm$0.3 & 79.3\tiny$\pm$1.0 & 85.5 \\
\midrule
\multirowcell{4}{\textit{Mixup-based} \\ \textit{Methods}} & Mixup \citep{wang2020heterogeneous} & 86.1\tiny$\pm$0.5 & 78.9\tiny$\pm$0.8 & \textbf{97.6}\tiny$\pm$0.1 & 75.8\tiny$\pm$1.8 & 84.6 \\
& Manifold Mixup\textsuperscript{\textdagger} \citep{verma2019manifold} & \textbf{88.8}\tiny$\pm$0.6 & 80.9\tiny$\pm$0.9 & 95.8\tiny$\pm$0.7 & 79.8\tiny$\pm$0.1 & 86.2 \\
& MixStyle\textsuperscript{\textdagger} \citep{zhou2020domain} & 83.7\tiny$\pm$0.1 & 80.7\tiny$\pm$3.7 & 95.5\tiny$\pm$1.2 & \textbf{81.4}\tiny$\pm$2.6 & 85.3 \\
& MetaMixup\textsuperscript{\textdagger} \citep{mai2021metamixup} & 84.9\tiny$\pm$1.5 & 79.6\tiny$\pm$0.6 & 96.3\tiny$\pm$0.9 & 80.1\tiny$\pm$1.4 & 85.2 \\
\midrule
\multirowcell{4}{\textit{Gradient-based} \\ \textit{Methods}} & PCGrad\textsuperscript{\textdagger} \citep{yu2020gradient} & 85.9\tiny$\pm$1.0 & 80.4\tiny$\pm$0.1 & 95.5\tiny$\pm$0.1 & 78.2\tiny$\pm$2.2 & 85.0 \\
& AND-Mask \citep{parascandolo2020learning} & 85.3\tiny$\pm$1.4 & 79.2\tiny$\pm$2.0 & 96.9\tiny$\pm$0.4 & 76.2\tiny$\pm$1.4 & 84.4\\
& Fish \citep{shi2021gradient} & - & - & - & - & 85.5 \\
& Fishr \citep{rame2021fishr} & 88.4\tiny$\pm$0.2 & 78.7\tiny$\pm$0.7 & 97.0\tiny$\pm$0.1 & 77.8\tiny$\pm$2.0 & 85.5 \\
\midrule
\multirowcell{3}{\textit{Augmentation} \\ \textit{Methods}} & L2A-OT\textsuperscript{\textdagger} \citep{zhou2020learning} & 87.9\tiny$\pm$1.5 & 81.4\tiny$\pm$2.3 & 96.7\tiny$\pm$1.3 & 77.2\tiny$\pm$2.4 & 85.8\\
& CNSN\textsuperscript{\textdagger} \citep{tang2021crossnorm} & 86.7\tiny$\pm$0.2 & 80.2\tiny$\pm$1.1 & 96.2\tiny$\pm$0.7 & 79.2\tiny$\pm$1.5 & 85.6\\
& DDG\textsuperscript{\textdagger} \citep{zhang2022towards} & 87.4\tiny$\pm$1.2 & 79.2\tiny$\pm$1.8 & 97.4\tiny$\pm$1.2 & 77.3\tiny$\pm$2.1 & 85.3\\
\midrule
\multirowcell{3}{\textit{DomainBed} \\ \textit{SOTA}}  & SelfReg \citep{kim2021selfreg} & 87.9\tiny$\pm$1.0 & 79.4\tiny$\pm$1.4 & 96.8\tiny$\pm$0.7 & 78.3\tiny$\pm$1.2 & 85.6 \\
& SagNet \citep{nam2021reducing} & 87.4\tiny$\pm$1.0 & 80.7\tiny$\pm$0.6 & 97.1\tiny$\pm$0.1 & 80.0\tiny$\pm$0.4 & 86.3 \\
& CORAL \citep{sun2016deep} & 88.3\tiny$\pm$0.2 & 80.0\tiny$\pm$0.5 & 97.5\tiny$\pm$0.3 & 78.8\tiny$\pm$1.3 & 86.2 \\ 
\midrule
& FGMix (ours) & 87.2\tiny$\pm$1.3 & \textbf{81.8}\tiny$\pm$1.2 & 97.3\tiny$\pm$0.7 & 80.7\tiny$\pm$1.6 & \textbf{86.8}\\
\midrule
\multicolumn{4}{c}{\textit{Combined with flatness-aware solver SWA \citep{izmailov2018averaging,arpit2021ensemble}}}  \\
\midrule
& ERM + SWA\textsuperscript{\textdagger} & 88.1\tiny$\pm$0.6 & 82.0\tiny$\pm$0.6 & 96.9\tiny$\pm$0.5 & 80.8\tiny$\pm$0.2 & 87.0 \\
& SelfReg + SWA & 85.9\tiny$\pm$0.6 & 81.9\tiny$\pm$0.4 & 96.8\tiny$\pm$0.1 & 81.4\tiny$\pm$0.6 & 86.5 \\
& CORAL + SWA\textsuperscript{\textdagger} & 87.6\tiny$\pm$0.3 & 83.1\tiny$\pm$0.7 & \textbf{97.5}\tiny$\pm$0.2 & 81.6\tiny$\pm$0.8 & 87.5 \\
& FGMix + SWA (ours) & \textbf{89.6}\tiny$\pm$0.7 & \textbf{83.8}\tiny$\pm$0.5 & 97.2\tiny$\pm$0.5 & \textbf{83.0}\tiny$\pm$0.8 & \textbf{88.4} \\
\bottomrule
\end{tabular}
%\vspace{-4mm}
\end{table}
\clearpage

\setlength{\tabcolsep}{3pt}
\begin{table}[!h]
\caption{Overall comparison of selected algorithms on \texttt{VLCS}.}
\label{tbl:overall_vlcs}
\scriptsize
\centering
\begin{tabular}{c l | c c c c  | c}
\toprule
& \textbf{Algorithm} & \texttt{C} & \texttt{L} & \texttt{S} & \texttt{V} & Avg. \\
\midrule
& ERM \citep{vapnick1998statistical} & 97.7\tiny$\pm$0.4 & 64.3\tiny$\pm$0.9 & 73.4\tiny$\pm$0.5 & 74.6\tiny$\pm$1.3 & 77.5 \\
\midrule
\multirowcell{4}{\textit{Mixup-based} \\ \textit{Methods}} & Mixup \citep{wang2020heterogeneous} & 98.3\tiny$\pm$0.6 & 64.8\tiny$\pm$1.0 & 72.1\tiny$\pm$0.5 & 74.3\tiny$\pm$0.8 & 77.4 \\
& Manifold Mixup\textsuperscript{\textdagger} \citep{verma2019manifold} & 97.4\tiny$\pm$1.3 & 63.8\tiny$\pm$1.9 & \textbf{73.7}\tiny$\pm$0.0 & 72.0\tiny$\pm$1.3 & 76.7 \\
& MixStyle\textsuperscript{\textdagger} \citep{zhou2020domain} & 97.9\tiny$\pm$0.7 & 63.3\tiny$\pm$1.7 & 70.7\tiny$\pm$0.2 & \textbf{77.5}\tiny$\pm$0.4 & 77.4 \\
& MetaMixup\textsuperscript{\textdagger} \citep{mai2021metamixup} & 98.6\tiny$\pm$1.2 & 63.4\tiny$\pm$0.8 & 72.8\tiny$\pm$0.1 & 76.7\tiny$\pm$1.1 & 77.9 \\
\midrule
\multirowcell{4}{\textit{Gradient-based} \\ \textit{Methods}} & PCGrad\textsuperscript{\textdagger} \citep{yu2020gradient} & 98.1\tiny$\pm$0.1 & \textbf{66.2}\tiny$\pm$0.1 & 70.1\tiny$\pm$1.1 & 75.7\tiny$\pm$1.9 & 77.5 \\
& AND-Mask \citep{parascandolo2020learning} & 97.8\tiny$\pm$0.4 & 64.3\tiny$\pm$1.2 & 73.5\tiny$\pm$0.7 & 76.8\tiny$\pm$2.6 & 78.1\\
& Fish \citep{shi2021gradient} & - & - & - & - & 77.8 \\
& Fishr \citep{rame2021fishr} & \textbf{98.9}\tiny$\pm$0.3 & 64.0\tiny$\pm$0.5 & 71.5\tiny$\pm$0.2 & 76.8\tiny$\pm$0.7 & 77.8 \\
\midrule
\multirowcell{3}{\textit{Augmentation} \\ \textit{Methods}} & L2A-OT\textsuperscript{\textdagger} \citep{zhou2020learning} & 98.2\tiny$\pm$0.8 & 64.1\tiny$\pm$0.4 & 71.6\tiny$\pm$1.5 & 75.5\tiny$\pm$0.8 & 77.4\\
& CNSN\textsuperscript{\textdagger} \citep{tang2021crossnorm} & 97.9\tiny$\pm$0.4 & 65.2\tiny$\pm$1.3 & 69.8\tiny$\pm$1.4 & 75.6\tiny$\pm$0.7 & 77.1\\
& DDG\textsuperscript{\textdagger} \citep{zhang2022towards} & 98.5\tiny$\pm$0.9 & 64.2\tiny$\pm$1.7 & 70.2\tiny$\pm$0.8 & 74.3\tiny$\pm$1.4 & 76.8 \\
\midrule
\multirowcell{3}{\textit{DomainBed} \\ \textit{SOTA}}  & SelfReg \citep{kim2021selfreg} & 96.7\tiny$\pm$0.4 & 65.2\tiny$\pm$1.2 & 73.1\tiny$\pm$1.3 & 76.2\tiny$\pm$0.7 & 77.8 \\
& SagNet \citep{nam2021reducing} & 97.9\tiny$\pm$0.4 & 64.5\tiny$\pm$0.5 & 71.4\tiny$\pm$1.3 & \textbf{77.5}\tiny$\pm$0.5 & 77.8 \\
& CORAL \citep{sun2016deep} & 98.3\tiny$\pm$0.1 & 66.1\tiny$\pm$1.2 & 73.4\tiny$\pm$0.3 & \textbf{77.5}\tiny$\pm$1.2 & \textbf{78.8} \\ 
\midrule
& FGMix (ours) & 98.1\tiny$\pm$0.4 & 63.9\tiny$\pm$0.9 & 73.1\tiny$\pm$1.4 & 77.1\tiny$\pm$0.8 & 78.1\\
\midrule
\multicolumn{4}{c}{\textit{Combined with flatness-aware solver SWA \citep{izmailov2018averaging,arpit2021ensemble}}}  \\
\midrule
& ERM + SWA\textsuperscript{\textdagger} & 97.0\tiny$\pm$0.9 & \textbf{64.3}\tiny$\pm$0.8 & 72.4\tiny$\pm$0.6 & 75.2\tiny$\pm$0.1 & 77.2 \\
& SelfReg + SWA & 97.4\tiny$\pm$0.4 & 63.5\tiny$\pm$0.3 & 72.6\tiny$\pm$0.1 & 76.7\tiny$\pm$0.7 & 77.5 \\
& CORAL + SWA\textsuperscript{\textdagger} & \textbf{98.6}\tiny$\pm$0.3 & 63.2\tiny$\pm$0.2 & 72.8\tiny$\pm$0.2 & \textbf{78.2}\tiny$\pm$1.1 & 78.2 \\
& FGMix + SWA (ours) & \textbf{98.6}\tiny$\pm$0.4 & 63.5\tiny$\pm$0.1 & \textbf{74.7}\tiny$\pm$0.6 & 78.1\tiny$\pm$1.2 & \textbf{78.7} \\
\bottomrule
\end{tabular}
%\vspace{-4mm}
\end{table}
\clearpage

\newpage
\begin{table}[!h]
\caption{Overall comparison of selected algorithms on \texttt{OfficeHome}.}
\label{tbl:overall_office}
\scriptsize
\centering
\begin{tabular}{c l | c c c c  | c}
\toprule
& \textbf{Algorithm} & \texttt{A} & \texttt{C} & \texttt{P} & \texttt{R} & Avg. \\
\midrule
& ERM \citep{vapnick1998statistical} & 61.3\tiny$\pm$0.7 & 52.4\tiny$\pm$0.3 & 75.8\tiny$\pm$0.1 & 76.6\tiny$\pm$0.3 & 66.5 \\
\midrule
\multirowcell{4}{\textit{Mixup-based} \\ \textit{Methods}} & Mixup \citep{wang2020heterogeneous} & 62.4\tiny$\pm$0.8 & 54.8\tiny$\pm$0.6 & \textbf{76.9}\tiny$\pm$0.3 & 78.3\tiny$\pm$0.2 & 68.1 \\
& Manifold Mixup\textsuperscript{\textdagger} \citep{verma2019manifold} & 61.6\tiny$\pm$0.8 & \textbf{55.1}\tiny$\pm$3.1 & 75.6\tiny$\pm$0.2 & 77.0\tiny$\pm$0.0 & 67.3 \\
& MixStyle\textsuperscript{\textdagger} \citep{zhou2020domain} & 61.7\tiny$\pm$0.6 & 53.3\tiny$\pm$1.6 & 76.3\tiny$\pm$1.0 & 77.8\tiny$\pm$0.5 & 67.3 \\
& MetaMixup\textsuperscript{\textdagger} \citep{mai2021metamixup} & 63.5\tiny$\pm$1.1 & 54.6\tiny$\pm$0.4 & 75.9\tiny$\pm$0.9 & 78.7\tiny$\pm$0.4 & 68.2 \\
\midrule
\multirowcell{4}{\textit{Gradient-based} \\ \textit{Methods}} & PCGrad\textsuperscript{\textdagger} \citep{yu2020gradient} & 60.6\tiny$\pm$1.1 & 52.1\tiny$\pm$1.7 & 74.4\tiny$\pm$0.6 & 74.0\tiny$\pm$0.3 & 65.5 \\
& AND-Mask \citep{parascandolo2020learning} & 59.5\tiny$\pm$1.2 & 51.7\tiny$\pm$0.2 & 73.9\tiny$\pm$0.4 & 77.1\tiny$\pm$0.2 & 65.6\\
& Fish \citep{shi2021gradient} & - & - & - & - & 68.6 \\
& Fishr \citep{rame2021fishr} & 62.4\tiny$\pm$0.5 & 54.4\tiny$\pm$0.4 & 76.2\tiny$\pm$0.5 & 78.3\tiny$\pm$0.1 & 67.8 \\
\midrule
\multirowcell{3}{\textit{Augmentation} \\ \textit{Methods}} & L2A-OT\textsuperscript{\textdagger} \citep{zhou2020learning} & 64.5\tiny$\pm$1.0 & 53.8\tiny$\pm$2.8 & 76.2\tiny$\pm$1.3 & 77.9\tiny$\pm$1.1 & 68.1 \\
& CNSN\textsuperscript{\textdagger} \citep{tang2021crossnorm} & 63.1\tiny$\pm$0.7 & 53.2\tiny$\pm$2.1 & 74.1\tiny$\pm$1.3 & 78.6\tiny$\pm$0.6 & 67.3\\
& DDG\textsuperscript{\textdagger} \citep{zhang2022towards} & 63.7\tiny$\pm$0.8 & 54.5\tiny$\pm$1.2 & 75.9\tiny$\pm$1.0 & 78.2\tiny$\pm$0.9 & 68.1 \\
\midrule
\multirowcell{3}{\textit{DomainBed} \\ \textit{SOTA}}  & SelfReg \citep{kim2021selfreg} & 63.6\tiny$\pm$1.4 & 53.1\tiny$\pm$1.0 & \textbf{76.9}\tiny$\pm$0.4 & 78.1\tiny$\pm$0.4 & 67.9 \\
& SagNet \citep{nam2021reducing} & 63.4\tiny$\pm$0.2 & 54.8\tiny$\pm$0.4 & 75.8\tiny$\pm$0.4 & 78.3\tiny$\pm$0.3 & 68.1 \\
& CORAL \citep{sun2016deep} & \textbf{65.3}\tiny$\pm$0.4 & 54.4\tiny$\pm$0.5 & 76.5\tiny$\pm$0.1 & 78.4\tiny$\pm$0.5 & 68.7 \\ 
\midrule
& FGMix (ours) & 65.2\tiny$\pm$1.5 & 54.6\tiny$\pm$2.1 & \textbf{76.9}\tiny$\pm$0.8 & \textbf{79.2}\tiny$\pm$0.8 & \textbf{69.0}\\
\midrule
\multicolumn{4}{c}{\textit{Combined with flatness-aware solver SWA \citep{izmailov2018averaging,arpit2021ensemble}}}  \\
\midrule
& ERM + SWA\textsuperscript{\textdagger} & 65.7\tiny$\pm$1.1 & 56.2\tiny$\pm$0.4 & 77.3\tiny$\pm$0.1 & 78.9\tiny$\pm$0.6 & 69.5 \\
& SelfReg + SWA & 64.9\tiny$\pm$0.8 & 55.4\tiny$\pm$0.6 & 78.4\tiny$\pm$0.2 & 78.8\tiny$\pm$0.1 & 69.4 \\
& CORAL + SWA\textsuperscript{\textdagger} & 68.3\tiny$\pm$0.2 & 57.0\tiny$\pm$0.0 & 77.9\tiny$\pm$0.3 & 79.7\tiny$\pm$0.0 & 70.7 \\
& FGMix + SWA (ours) & \textbf{68.5}\tiny$\pm$1.0 & \textbf{57.6}\tiny$\pm$0.8 & \textbf{78.8}\tiny$\pm$0.5 & \textbf{80.4}\tiny$\pm$0.2 & \textbf{71.3} \\
\bottomrule
\end{tabular}
%\vspace{-4mm}
\end{table}
\clearpage

\begin{table}[!h]
\caption{Overall comparison of selected algorithms on \texttt{TerraIncognita}.}
\label{tbl:overall_terra}
\scriptsize
\centering
\begin{tabular}{c l | c c c c  | c}
\toprule
& \textbf{Algorithm} & \texttt{L100} & \texttt{L38} & \texttt{L43} & \texttt{L46} & Avg. \\
\midrule
& ERM \citep{vapnick1998statistical} & 49.8\tiny$\pm$4.4 & 42.1\tiny$\pm$1.4 & 56.9\tiny$\pm$1.8 & 35.7\tiny$\pm$3.9 & 46.1 \\
\midrule
\multirowcell{4}{\textit{Mixup-based} \\ \textit{Methods}} & Mixup \citep{wang2020heterogeneous} & \textbf{59.6}\tiny$\pm$2.0 & 42.2\tiny$\pm$1.4 & 55.9\tiny$\pm$0.8 & 33.9\tiny$\pm$1.4 & 47.9 \\
& Manifold Mixup\textsuperscript{\textdagger} \citep{verma2019manifold} & 57.7\tiny$\pm$2.9 & 42.4\tiny$\pm$3.0 & 55.8\tiny$\pm$0.4 & 39.2\tiny$\pm$1.9 & 48.8 \\
& MixStyle\textsuperscript{\textdagger} \citep{zhou2020domain} & 53.9\tiny$\pm$0.6 & 42.3\tiny$\pm$0.9 & 53.2\tiny$\pm$2.7 & 37.6\tiny$\pm$0.1 & 46.8 \\
& MetaMixup\textsuperscript{\textdagger} \citep{mai2021metamixup} & 53.2\tiny$\pm$1.7 & 43.7\tiny$\pm$2.6 & 54.5\tiny$\pm$1.9 & 36.9\tiny$\pm$1.1 & 47.1 \\
\midrule
\multirowcell{4}{\textit{Gradient-based} \\ \textit{Methods}} & PCGrad\textsuperscript{\textdagger} \citep{yu2020gradient} & 55.9\tiny$\pm$2.0 & 43.2\tiny$\pm$3.8 & 56.4\tiny$\pm$2.5 & 37.7\tiny$\pm$1.0 & 48.3 \\
& AND-Mask \citep{parascandolo2020learning} & 50.0\tiny$\pm$2.9 & 40.2\tiny$\pm$0.8 & 53.3\tiny$\pm$0.7 & 34.8\tiny$\pm$1.9 & 44.6\\
& Fish \citep{shi2021gradient} & - & - & - & - & 45.1 \\
& Fishr \citep{rame2021fishr} & 50.2\tiny$\pm$3.9 & 43.9\tiny$\pm$0.8 & 55.7\tiny$\pm$2.2 & 39.8\tiny$\pm$1.0 & 47.4 \\
\midrule
\multirowcell{3}{\textit{Augmentation} \\ \textit{Methods}} & L2A-OT\textsuperscript{\textdagger} \citep{zhou2020learning} & 58.7\tiny$\pm$2.0 & 43.5\tiny$\pm$2.7 & 54.8\tiny$\pm$1.9 & 37.4\tiny$\pm$1.8 & 48.6\\
& CNSN\textsuperscript{\textdagger} \citep{tang2021crossnorm} & 58.2\tiny$\pm$1.9 & 43.2\tiny$\pm$2.1 & 57.1\tiny$\pm$1.3 & 35.2\tiny$\pm$1.1 & 48.4\\
& DDG\textsuperscript{\textdagger} \citep{zhang2022towards} & \textbf{59.6}\tiny$\pm$3.0 & 41.2\tiny$\pm$2.4 & 56.0\tiny$\pm$1.8 & 33.9\tiny$\pm$1.2 & 47.7\\
\midrule
\multirowcell{3}{\textit{DomainBed} \\ \textit{SOTA}}  & SelfReg \citep{kim2021selfreg} & 48.8\tiny$\pm$0.9 & 41.3\tiny$\pm$1.8 & 57.3\tiny$\pm$0.7 & 40.6\tiny$\pm$0.9 & 47.0 \\
& SagNet \citep{nam2021reducing} & 53.0\tiny$\pm$2.9 & 43.0\tiny$\pm$2.5 & \textbf{57.9}\tiny$\pm$0.6 & 40.4\tiny$\pm$1.3 & 48.6 \\
& CORAL \citep{sun2016deep} & 51.6\tiny$\pm$2.4 & 42.2\tiny$\pm$1.0 & 57.0\tiny$\pm$1.0 & 39.8\tiny$\pm$2.9 & 47.6 \\ 
\midrule
& FGMix (ours) & 54.1\tiny$\pm$2.7 & \textbf{44.7}\tiny$\pm$1.1 & 57.0\tiny$\pm$1.8 & \textbf{40.9}\tiny$\pm$0.8 & \textbf{49.2}\\
\midrule
\multicolumn{4}{c}{\textit{Combined with flatness-aware solver SWA \citep{izmailov2018averaging,arpit2021ensemble}}}  \\
\midrule
& ERM + SWA\textsuperscript{\textdagger} & 53.5\tiny$\pm$0.9 & 47.6\tiny$\pm$0.8 & 58.2\tiny$\pm$0.4 & 41.0\tiny$\pm$0.8 & 50.1 \\
& SelfReg + SWA & 56.8\tiny$\pm$0.9 & 44.7\tiny$\pm$0.6 & 59.6\tiny$\pm$0.3 & \textbf{42.9}\tiny$\pm$0.8 & 51.0 \\
& CORAL + SWA\textsuperscript{\textdagger} & 55.6\tiny$\pm$0.5 & 48.1\tiny$\pm$0.7 & 58.5\tiny$\pm$0.1 & 42.2\tiny$\pm$1.0 & 51.1 \\
& FGMix + SWA (ours) & \textbf{57.1}\tiny$\pm$0.8 & \textbf{49.5}\tiny$\pm$1.3 & \textbf{60.7}\tiny$\pm$0.3 & 41.8\tiny$\pm$1.1 & \textbf{52.3} \\
\bottomrule
\end{tabular}
%\vspace{-4mm}
\end{table}
\clearpage

\begin{table}[!h]
\caption{Overall comparison of selected algorithms on \texttt{DomainNet}.}
\label{tbl:overall_net}
\scriptsize
\centering
\begin{tabular}{c l | c c c c c c | c}
\toprule
& \textbf{Algorithm} & \texttt{clip} & \texttt{info} & \texttt{paint} & \texttt{quick} & \texttt{real} & \texttt{sketch} & Avg. \\
\midrule
& ERM \citep{vapnick1998statistical} & 58.1\tiny$\pm$0.3 & 18.8\tiny$\pm$0.3 & 46.7\tiny$\pm$0.3 & 12.2\tiny$\pm$0.4 & 59.6\tiny$\pm$0.1 & 49.8\tiny$\pm$0.4 & 40.9 \\
\midrule
\multirowcell{4}{\textit{Mixup-based} \\ \textit{Methods}} & Mixup \citep{wang2020heterogeneous} & 55.7\tiny$\pm$0.3 & 18.5\tiny$\pm$0.5 & 44.3\tiny$\pm$0.5 & 12.5\tiny$\pm$0.4 & 55.8\tiny$\pm$0.3 & 48.2\tiny$\pm$0.5 & 39.2 \\
& Manifold Mixup\textsuperscript{\textdagger} \citep{verma2019manifold} & 60.7\tiny$\pm$0.4 & 19.4\tiny$\pm$0.1 & 47.1\tiny$\pm$0.3 & 11.4\tiny$\pm$0.2 & 59.6\tiny$\pm$0.6 & 48.7\tiny$\pm$0.2 & 41.2 \\
& MixStyle\textsuperscript{\textdagger} \citep{zhou2020domain} & 59.9\tiny$\pm$0.2 & 19.0\tiny$\pm$0.3 & 47.0\tiny$\pm$0.1 & 11.5\tiny$\pm$0.1 & 58.9\tiny$\pm$0.4 & 48.8\tiny$\pm$0.0 & 40.9 \\
& MetaMixup\textsuperscript{\textdagger} \citep{mai2021metamixup} & 60.7\tiny$\pm$0.3 & 20.0\tiny$\pm$0.5 & 47.1\tiny$\pm$0.4 & 12.8\tiny$\pm$0.1 & 60.1\tiny$\pm$0.1 & 50.1\tiny$\pm$0.3 & 41.8 \\
\midrule
\multirowcell{4}{\textit{Gradient-based} \\ \textit{Methods}} & PCGrad\textsuperscript{\textdagger} \citep{yu2020gradient} & 60.3\tiny$\pm$0.2 & 18.1\tiny$\pm$0.4 & 47.0\tiny$\pm$0.4 & 12.9\tiny$\pm$0.1 & 59.8\tiny$\pm$0.0 & 48.4\tiny$\pm$0.3 & 41.1 \\
& AND-Mask \citep{parascandolo2020learning} & 52.3\tiny$\pm$0.8 & 16.6\tiny$\pm$0.3 & 41.6\tiny$\pm$1.1 & 11.3\tiny$\pm$0.1 & 55.8\tiny$\pm$0.4 & 45.4\tiny$\pm$0.9 & 37.2\\
& Fish \citep{shi2021gradient} & - & - & - & - & - & - & \textbf{42.7} \\
& Fishr \citep{rame2021fishr} & 58.2\tiny$\pm$0.5 & 20.2\tiny$\pm$0.2 & \textbf{47.7}\tiny$\pm$0.3 & 12.7\tiny$\pm$0.2 & 60.3\tiny$\pm$0.2 & \textbf{50.8}\tiny$\pm$0.1 & 41.7 \\
\midrule
\multirowcell{3}{\textit{Augmentation} \\ \textit{Methods}} & L2A-OT\textsuperscript{\textdagger} \citep{zhou2020learning} & 58.7\tiny$\pm$0.3 & 18.5\tiny$\pm$0.4 & 46.2\tiny$\pm$0.5 & 11.3\tiny$\pm$0.7 & 57.6\tiny$\pm$0.8 & 48.9\tiny$\pm$0.2 & 40.2\\
& CNSN\textsuperscript{\textdagger} \citep{tang2021crossnorm} & 60.2\tiny$\pm$0.2 & 19.0\tiny$\pm$0.1 & 46.3\tiny$\pm$0.5 & 11.6\tiny$\pm$0.3 & 58.2\tiny$\pm$0.6 & 48.7\tiny$\pm$0.4 & 40.7\\
& DDG\textsuperscript{\textdagger} \citep{zhang2022towards} & 59.7\tiny$\pm$0.5 & 18.7\tiny$\pm$0.2 & 45.1\tiny$\pm$0.9 & 12.1\tiny$\pm$0.4 & 56.8\tiny$\pm$0.3 & 47.5\tiny$\pm$0.7 & 40.0\\
\midrule
\multirowcell{3}{\textit{DomainBed} \\ \textit{SOTA}}  & SelfReg \citep{kim2021selfreg} & 58.5\tiny$\pm$0.1 & \textbf{20.7}\tiny$\pm$0.1 & 47.3\tiny$\pm$0.3 & 13.1\tiny$\pm$0.3 & 58.2\tiny$\pm$0.2 & 51.1\tiny$\pm$0.3 & 41.5 \\
& SagNet \citep{nam2021reducing} & 57.7\tiny$\pm$0.3 & 19.0\tiny$\pm$0.2 & 45.3\tiny$\pm$0.3 & 12.7\tiny$\pm$0.5 & 58.1\tiny$\pm$0.5 & 48.8\tiny$\pm$0.2 & 40.3 \\
& CORAL \citep{sun2016deep} & 59.2\tiny$\pm$0.1 & 19.7\tiny$\pm$0.2 & 46.6\tiny$\pm$0.3 & \textbf{13.4}\tiny$\pm$0.4 & 59.8\tiny$\pm$0.2 & 50.1\tiny$\pm$0.6 & 41.5 \\ 
\midrule
& FGMix (ours) & \textbf{61.1}\tiny$\pm$0.1 & 20.4\tiny$\pm$0.6 & 47.0\tiny$\pm$0.6 & 12.5\tiny$\pm$0.5 & \textbf{61.1}\tiny$\pm$0.8 & 50.7\tiny$\pm$0.3 & 42.1\\
\midrule
\multicolumn{4}{c}{\textit{Combined with flatness-aware solver SWA \citep{izmailov2018averaging,arpit2021ensemble}}}  \\
\midrule
& ERM + SWA\textsuperscript{\textdagger} & 62.0\tiny$\pm$0.1 & 21.0\tiny$\pm$0.0 & 50.8\tiny$\pm$0.2 & 13.6\tiny$\pm$0.3 & 62.7\tiny$\pm$0.1 & \textbf{54.0}\tiny$\pm$0.2 & 44.0 \\
& SelfReg + SWA & 62.4\tiny$\pm$0.1 & \textbf{22.6}\tiny$\pm$0.1 & 51.8\tiny$\pm$0.1 & 14.3\tiny$\pm$0.1 & 62.5\tiny$\pm$0.2 & 53.8\tiny$\pm$0.3 & 44.6 \\
& CORAL + SWA\textsuperscript{\textdagger} & 63.0\tiny$\pm$0.3 & 22.1\tiny$\pm$0.5 & 51.7\tiny$\pm$0.4 & 14.9\tiny$\pm$0.5 & 63.0\tiny$\pm$0.3 & 53.1\tiny$\pm$0.1 & 44.6 \\
& FGMix + SWA (ours) & \textbf{63.1}\tiny$\pm$0.2 & 22.3\tiny$\pm$0.7 & \textbf{52.2}\tiny$\pm$0.5 & \textbf{15.1}\tiny$\pm$0.2 & \textbf{63.9}\tiny$\pm$0.2 & \textbf{54.0}\tiny$\pm$0.3 & \textbf{45.1} \\
\bottomrule
\end{tabular}
%\vspace{-4mm}
\end{table}
\clearpage

\newpage
\subsection{Full Results for Ablation Study}
\label{appx:ablation}

\setlength{\tabcolsep}{2.5pt}
%\vspace{-2mm}
\begin{table}[!h]
\caption{Ablation study of FGMix on \texttt{PACS}.}
\label{tbl:ablation_pacs}
\scriptsize
\centering
\begin{tabular}{c c c c c c c c c c c}
%\begin{tabular}{|*{11}{c|}}
\toprule
\multirowcell{2}{\textbf{Variant}}
& \multirowcell{2}{\scriptsize similarity-\\ \scriptsize based weights} & \multirowcell{2}{\scriptsize gradient-\\ \scriptsize based similarity} & \multirowcell{2}{\scriptsize scaling \\ \scriptsize \& shifting} & \multirowcell{2}{\scriptsize$\mathcal{L}_{flat}$} & \multirow{2}{*}{\scriptsize$\mathcal{L}_{adv}$} & \multirow{2}{*}{\texttt{A}} & \multirow{2}{*}{\texttt{C}} & \multirow{2}{*}{\texttt{P}} & \multirow{2}{*}{\texttt{S}} & \multirow{2}{*}{Avg.}\\
\\
\midrule
A (baseline) &&&&&& 84.3\tiny$\pm$0.8 & 80.2\tiny$\pm$0.7 & 94.8\tiny$\pm$0.3 & 79.6\tiny$\pm$1.0 & 84.7\\
\midrule
B & \checkmark &&&&& 85.1\tiny$\pm$0.4 & 80.9\tiny$\pm$0.8 & 95.3\tiny$\pm$0.2 & 79.6\tiny$\pm$0.6 & 85.2 \\
C & \checkmark & \checkmark &&&& 86.6\tiny$\pm$0.2 & 80.1\tiny$\pm$0.5 & 95.9\tiny$\pm$0.1 & 79.7\tiny$\pm$0.1 & 85.6 \\
D & \checkmark & \checkmark & \checkmark &&& 87.5\tiny$\pm$1.5 & 81.0\tiny$\pm$1.4 & 95.6\tiny$\pm$0.7 & 80.0\tiny$\pm$0.7 & 86.0 \\
E & \checkmark & \checkmark & \checkmark & \checkmark && \textbf{88.0}\tiny$\pm$1.7 & 81.5\tiny$\pm$2.3 & 96.9\tiny$\pm$1.4 & 80.3\tiny$\pm$1.4 & 86.7\\
F & \checkmark &  & \checkmark & \checkmark &  \checkmark & 87.9\tiny$\pm$1.2 & 80.9\tiny$\pm$1.9 & 96.1\tiny$\pm$1.3 & 80.0\tiny$\pm$1.6 & 86.2 \\ 
\midrule
FGMix (ours) & \checkmark & \checkmark & \checkmark & \checkmark & \checkmark & 87.2\tiny$\pm$1.3 & \textbf{81.8}\tiny$\pm$1.2 & \textbf{97.3}\tiny$\pm$0.7 & \textbf{80.7}\tiny$\pm$1.6 & \textbf{86.8} \\
\bottomrule
\end{tabular}
%\vspace{-2mm}
\end{table}

\setlength{\tabcolsep}{2.5pt}
%\vspace{-2mm}
\begin{table}[!h]
\caption{Ablation study of FGMix on \texttt{TerraIncognita}.}
\label{tbl:ablation_terra}
\scriptsize
\centering
\begin{tabular}{c c c c c c c c c c c}
%\begin{tabular}{|*{11}{c|}}
\toprule
\multirowcell{2}{\textbf{Variant}} & \multirowcell{2}{\scriptsize similarity-\\ \scriptsize based weights} & \multirowcell{2}{\scriptsize gradient-\\ \scriptsize based similarity} & \multirowcell{2}{\scriptsize scaling \\ \scriptsize \& shifting} & \multirowcell{2}{\scriptsize$\mathcal{L}_{flat}$} & \multirow{2}{*}{\scriptsize$\mathcal{L}_{adv}$} & \multirow{2}{*}{\texttt{L100}} & \multirow{2}{*}{\texttt{L38}} & \multirow{2}{*}{\texttt{L43}} & \multirow{2}{*}{\texttt{L46}} & \multirow{2}{*}{Avg.}\\
\\
\midrule
A (baseline) &&&&&& 51.1\tiny$\pm$0.7 & 43.2\tiny$\pm$1.4 & 56.4\tiny$\pm$1.2 & 39.3\tiny$\pm$1.1 & 47.5\\
\midrule
B & \checkmark &&&&& 50.4\tiny$\pm$1.0 & 43.9\tiny$\pm$1.2 & 57.0\tiny$\pm$0.8 & 40.1\tiny$\pm$0.5 & 47.9 \\
C & \checkmark & \checkmark &&&& 50.4\tiny$\pm$0.8 & \textbf{44.7}\tiny$\pm$1.6 & 57.5\tiny$\pm$0.6 & 39.9\tiny$\pm$1.3 & 48.1 \\
D & \checkmark & \checkmark & \checkmark &&& 51.9\tiny$\pm$1.7 & 44.2\tiny$\pm$2.0 & 58.0\tiny$\pm$1.1 & \textbf{41.1}\tiny$\pm$1.9 & 48.8 \\
E & \checkmark & \checkmark & \checkmark & \checkmark && 54.0\tiny$\pm$2.1 & 44.3\tiny$\pm$3.0 & \textbf{58.1}\tiny$\pm$1.9 & 40.5\tiny$\pm$1.8 & \textbf{49.2}\\
F & \checkmark &  & \checkmark & \checkmark &  \checkmark & 53.1\tiny$\pm$1.9 & 44.0\tiny$\pm$2.2 & 57.9\tiny$\pm$1.2 & 40.7\tiny$\pm$1.6 & 48.9 \\ 
\midrule
FGMix (ours) & \checkmark & \checkmark & \checkmark & \checkmark & \checkmark & \textbf{54.1}\tiny$\pm$1.7 & \textbf{44.7}\tiny$\pm$1.1 & 57.0\tiny$\pm$1.8 & 40.9\tiny$\pm$0.8 & \textbf{49.2} \\
\bottomrule
\end{tabular}
%\vspace{-2mm}
\end{table}

\newpage
\subsection{Sensitivity Tests of Varying Loss Weighting $\alpha$ and Initial $\lambda$}
\label{appx:sens}
Given \eqref{eqn:overall}, the loss weighting $\alpha$ basically determines how strongly the adversarial constraint should be applied to obtain a desirable mixup distribution. If $\alpha$ is too large, the training of the weight generation policy will collapse and tend to generate a distribution that closely mimics the source distribution. If $\alpha$ is too small, the constraint becomes ineffective, leading to unregularized extrapolation that deviates substantially from the reasonable region, thereby severely affecting base-model learning. Since the adversarial constraint prevents the detrimental over-extrapolation effect, the intensity of the training signal is related to the initialization of the scaling factor $\lambda$, which generally determines the extent of extrapolation.\ 
%For a small initial $\lambda$ (recall that $\lambda=1$ is equivalent to no extrapolation), the mixup data are mostly within the convex hull of source domains. In that case, assigning a large $\alpha$ for $\mathcal{L}_{adv}$ is less meaningful and may potentially degrade the performance, as the signal from $\mathcal{L}_{flat}$ is weakened instead. If the initial $\lambda$ is large, it is important to impose a stronger adversarial constraint, as failing to do so may result in generating less relevant, seriously drifted mixup data, which are detrimental for base model training. \

Table B.11 presents the sensitivity tests of varying $\alpha$ and initial $\lambda$ on model performance on the \texttt{TerraInc} dataset. We vary $\alpha$ in \{0.01, 0.03, 0.1, 0.3\} and initial $\lambda$ in \{1, 3, 5, 8\}. 
%(mentioned in Section 4.1, paragraph 3, page 13)
Firstly, we observe that when the initial $\lambda$ is set to 1 (no extrapolation in the beginning), a smaller $\alpha$ value results in better performance, as the lightly imposed adversarial constraint allows $\lambda$ to learn useful extrapolation from the flatness-aware loss more freely. When we increase the initial $\lambda$ to more than 1 (extrapolation exists from the beginning), we can see that a greater $\alpha$ is more beneficial, as the stronger adversarial constraint helps adjust mixup data that significantly drift from the source domains, which are detrimental for base model training. Overall, the model's performance is sensitive to $\lambda$. A large initial $\lambda$ may simply crash the base model training and results in poor performance. This also underscores the importance of applying the adversarial constraint, which draws the extrapolated data closer to the source domains, thereby mitigating the adverse effects of large $\lambda$. In this specific case, we found that a moderately low value for $\alpha$ (around 0.1 to 0.3) yields the best result when the initial $\lambda$ is 3.

\setlength{\tabcolsep}{6pt}
\begin{table}[ht]
\centering
\footnotesize
%\vspace{-4mm}
%\caption{Sensitivity tests of varying loss weighting $\alpha$ and initial $\lambda$ on \texttt{TerraInc} dataset.}
\caption{Sensitivity tests of varying loss weighting $\alpha$ and initial $\lambda$ on \texttt{TerraInc} dataset.}
%\caption{Your Caption Here}
\label{tab:hyper}
\begin{tabular}{ccccc}
\toprule
 & \multicolumn{4}{c}{$\alpha$ for $\mathcal{L}_{adv}$} \\
\cmidrule(lr){2-5}
initial $\lambda$ & 0.01 & 0.03 & 0.1 & 0.3 \\
\midrule
1 & 47.9 & 48.2 & 46.3 & 46.5 \\
3 & 47.3 & 47.1 & \textbf{49.2} & 49.0\\
5 & 44.3 & 44.0 & 46.5 & 46.1 \\
8 & 43.1 & 44.0 & 44.8 & 45.1\\
\bottomrule
\end{tabular}
\vspace{-2mm}
\end{table}

\newpage
\subsection{Effect of Mixup on the Learned Feature Representations}
\label{appx:vis}

Though FGMix performs mixup after the feature extractor $g_{\theta}$ to regularize the classifier, the gradients induced by mixup naturally propagate back into $g_{\theta}$ and can indirectly influence the learned features through end-to-end training.

\begin{figure}
\centering
\includegraphics[width=0.9\columnwidth]{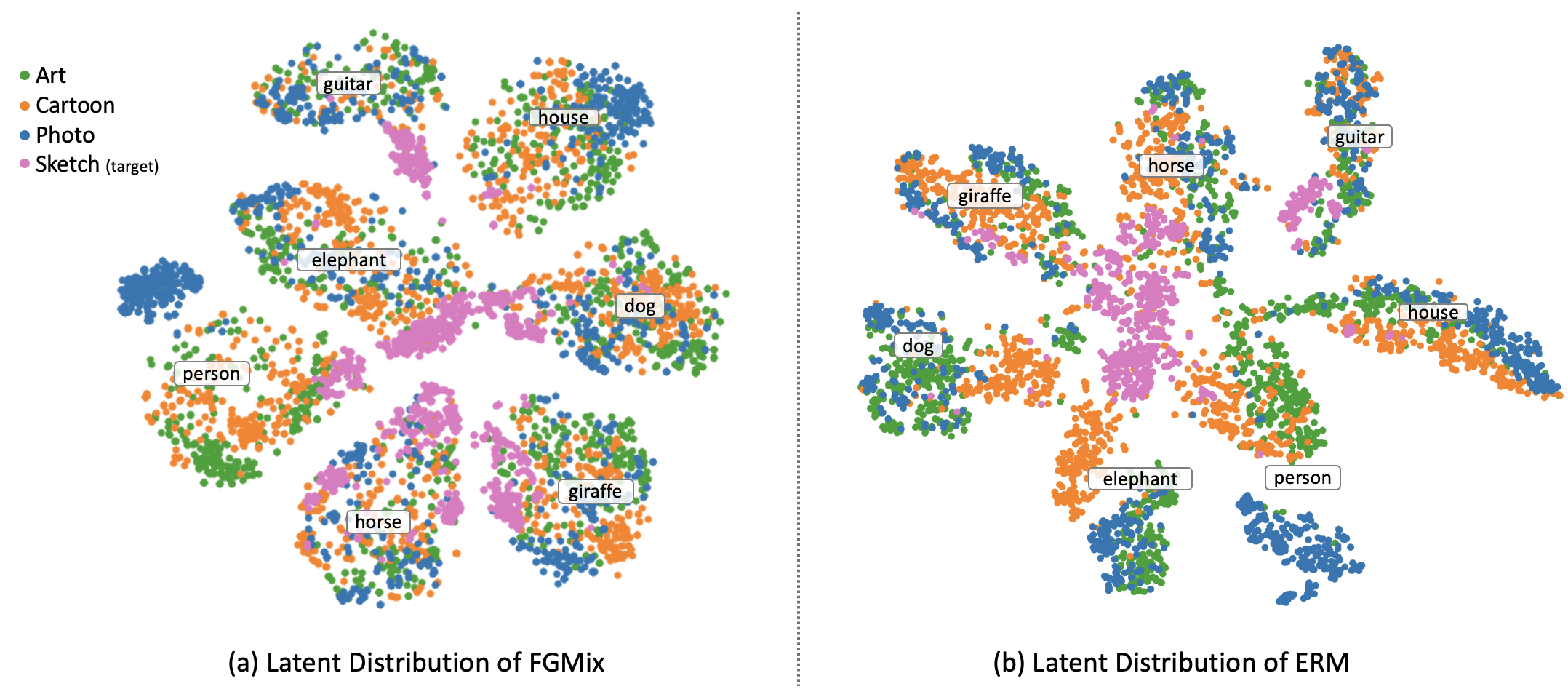}
\caption{t-SNE visualization of the latent distributions on \texttt{PACS} trained with (a) FGMix and (b) ERM. In both plots, we use ``\texttt{Art}", ``\texttt{Cartoon}", ``\texttt{Photo}" as the source domains and ``\texttt{Sketch}" as the target domain.}
\label{fig:feat_dist}
\end{figure}

To empirically examine the effect of mixup on the learned feature representations of feature extractor $g_{\theta}$, we visualize the latent distributions produced by $g_{\theta}$ under FGMix (with mixup) and ERM (without mixup) on the PACS dataset using t-SNE, as shown in Figure \ref{fig:feat_dist}. From the visualizations, several observations can be made: both ERM and FGMix learn well-separated class clusters, demonstrating that both methods produce a strong class-discriminative structure of the latent space; ERM exhibits more pronounced domain-specific subclusters within each class (e.g., “elephant” and “dog”), where samples from the \texttt{Cartoon} domain (orange) form a subcluster that is visibly segregated from the other source domains; in contrast, FGMix produces more cohesive, domain-mixed clusters in which samples from different domains are more evenly distributed within each class cluster, and the domain-specific structure is noticeably reduced. A possible explanation is that mixing latent codes from different domains within the same class encourages the classifier to rely less on domain-specific artifacts. As a result, the induced gradients promote the learning of features that emphasize class-relevant information while deemphasizing domain-induced variations. {This induced feature geometry may also facilitate the robust classification of extrapolated out-of-domain representations under parameter perturbations, and we leave a more thorough investigation of this effect to future work.}\

While this improvement in latent-space cohesion is a byproduct of mixup training, the main benefit of FGMix still lies in its ability to extrapolate beyond the convex hull of the source domains—an effect neither ERM nor interpolation-only mixup achieves (as shown previously in Figure \ref{fig:distribution}). This is also reflected in the t-SNE plots, where target-domain samples (\texttt{Sketch}, pink) remain entangled near the center along class boundaries, highlighting where extrapolated mixup instances are particularly helpful. 

\clearpage

\section{2D Loss/Accuracy Surface Construction}
\label{appx:loss_surface_const}
Following \citet{izmailov2018averaging,cha2021swad}, we select 3 model weights $\{w_1,w_2,w_3\}$ at the initialization, the end of ERM + SWA optimization path, and the end of FGMix + SWA optimization path, respectively. Note that $w_i$ is the concatenation of vectorized $\theta_i$ and $\phi_i$ for the $i$-th selected model. We define the orthogonal basis $\{u,v\}$ of the 2D plane as:
$$
u=w_2-w_1, \ \ \ \ v=\frac{(w_3-w_1)-\langle w_3-w_1,w_2-w_1\rangle}{\|w_2-w_1\|^2\cdot(w_2-w_1)}.
$$
The orthonormal basis of the 2D plane is $\hat{u}=u/\|u\|$ and $\hat{v}=v/\|v\|$. \

We project the optimization path onto the 2D plane by computing the coordinates of each point on the path. That is, for the point at the $j$-th step, its $u$- and $v$-coordinates are $\langle(w_j-w_1),\hat{u}\rangle$ and $\langle(w_j-w_1),\hat{v}\rangle$ respectively. To plot the loss surface, we define ranges on  $u$- and $v$-axes that fully contain the optimization paths, say $[u_1,u_2]$ for $u$-axis and $[v_1,v_2]$ for $v$-axis. We then obtain the model weight corresponding to each grid point in the defined range, i.e., $w=w_1+a\hat{u}+b\hat{v}$, $\forall a\in[u_1,u_2], b\in[v_1,v_2]$, and compute the loss/accuracy of the entire training/test dataset using the model weight. The loss/accuracy values are visualized with a contour plot. For direct comparison, we use the same $u,v$ ranges for all three plots (ERM \& FGMix train loss plots and test accuracy plot), and set the scale interval to be the same for the two train loss plots for flatness comparison.







% \end{thebibliography}
\end{document}